%% file: main.tex
\documentclass[sigconf]{acmart}

\usepackage{multirow}
\usepackage{colortbl}
\usepackage{enumerate}
\usepackage{amsmath}
\usepackage{array}
\usepackage{xspace}

\def \alambic {\includegraphics[width=0.01\linewidth]{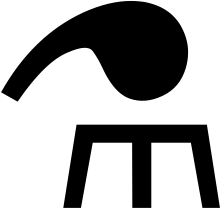}\xspace}

\AtBeginDocument{%
  }

\copyrightyear{2023}
\acmYear{2023}
\setcopyright{acmlicensed}
\acmConference[MM '23] {Proceedings of the 31st ACM International Conference on Multimedia}{October 29--November 3, 2023}{Ottawa, ON, Canada.}
\acmBooktitle{Proceedings of the 31st ACM International Conference on Multimedia (MM '23), October 29--November 3, 2023, Ottawa, ON, Canada}
\acmPrice{15.00}
\acmISBN{979-8-4007-0108-5/23/10}
\acmDOI{10.1145/3581783.3611762}

\begin{document}

\title{LGViT: Dynamic Early Exiting for Accelerating Vision Transformer}

\author{Guanyu Xu}
\authornote{Both authors contributed equally to this research.}
\affiliation{%
  \institution{Beijing Institute of Technology}
  \city{Beijing}
  \country{China}}
\email{xuguanyu@bit.edu.cn}

\author{Jiawei Hao}
\authornotemark[1]
\affiliation{%
  \institution{Beijing Institute of Technology}
  \city{Beijing}
  \country{China}}
\email{haojiawei7@bit.edu.cn}

\author{Li Shen}
\affiliation{%
  \institution{JD Explore Academy}
  \city{Beijing}
  \country{China}}
\email{mathshenli@gmail.com}

\author{Han Hu}
\authornote{Han Hu is the corresponding author.}
\affiliation{%
  \institution{Beijing Institute of Technology}
  \city{Beijing}
  \country{China}
  \postcode{100081}
}
\email{hhu@bit.edu.cn}

\author{Yong Luo}
\affiliation{%
  \institution{Wuhan University}
  \city{Wuhan}
  \country{China}
  \postcode{430072}
}
\email{luoyong@whu.edu.cn}

\author{Hui Lin}
\affiliation{%
  \institution{China Academic of Electronics and Information Technology}
  \city{Beijing}
  \country{China}
  \postcode{100041}
}
\email{linhui@whu.edu.cn}

\author{Jialie Shen}
\affiliation{%
  \institution{City, University of London}
  \city{London}
  \country{U.K.}
  \postcode{EC1V 0HB}
}
\email{jialie@gmail.com}

\renewcommand{\shortauthors}{Xu et al.}

\begin{abstract}
  Recently, the efficient deployment and acceleration of powerful vision transformers (ViTs) on resource-limited edge devices for providing multimedia services have become attractive tasks. 
  Although early exiting is a feasible solution for accelerating inference, most works focus on convolutional neural networks (CNNs) and transformer models in natural language processing (NLP). 
  Moreover, the direct application of early exiting methods to ViTs may result in substantial performance degradation. 
  To tackle this challenge, 
  we systematically investigate the efficacy of early exiting in ViTs and point out that the insufficient feature representations in shallow internal classifiers and the limited ability to capture target semantic information in deep internal classifiers restrict the performance of these methods. 
  We then propose an early exiting framework for general ViTs termed \textbf{LGViT}, which incorporates heterogeneous exiting heads, namely, 
  local perception head and global aggregation head, to achieve an efficiency-accuracy trade-off. 
  In particular, we develop a novel two-stage training scheme, including end-to-end training and self-distillation with the backbone frozen to generate early exiting ViTs, 
  which facilitates the fusion of global and local information extracted by the two types of heads. 
  We conduct extensive experiments using three popular ViT backbones on three vision datasets. 
  Results demonstrate that our LGViT can achieve competitive performance with approximately 1.8 $\times$ speed-up. 

\end{abstract}

\ccsdesc[500]{Computing methodologies~Computer vision tasks}

\keywords{Vision transformer, early exit, heterogeneous exiting heads, self-distillation}


\maketitle

  \input{sections/intro.tex}

  \input{sections/related_work.tex}

\input{sections/method.tex}

  \input{sections/experiment.tex}

  \input{sections/conclusion.tex}

\begin{acks}
  This work was partially supported by the National Key Research and Development Program of China under No. 2021YFC3300200 and the National Natural Science Foundation of China under Grants No. 61971457.
\end{acks}

\bibliographystyle{ACM-Reference-Format}
{\balance \bibliography{ref}}

\appendix

\input{sections/appendix.tex}

\end{document}

%% file: sections/intro.tex
\section{Introduction}
\label{sc:intro}

\begin{figure}
    \begin{center}
      \includegraphics[width=0.42\textwidth]{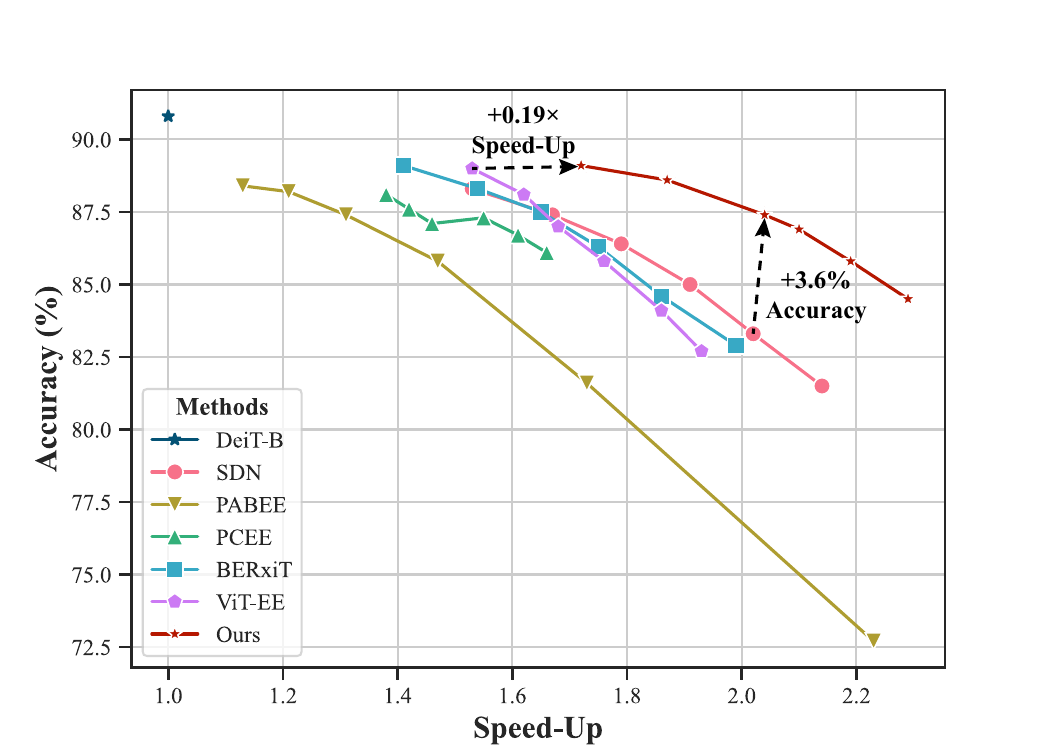}
    \end{center}
    \vspace{-8pt}
    \caption{The comparison of performance and efficiency trade-off for the ViT backbone in CIFAR-100. LGViT significantly outperforms other early exiting methods. 
    In particular, LGViT achieves new state-of-the-art 89.1 \% accuracy but faster than ViT-EE \cite{ViT-EE}. Details are in Table \ref{tb:performance} and Section \ref{sc:peformance}. }
    \Description{Trade-off.}
    \vspace{-12pt}
    \label{fig:trade-off}
  \end{figure}

During the past few years, vision transformers (ViTs) have become fundamental backbones for various multimedia tasks due to their powerful performance and universal structures \cite{vit, DBLP:conf/mm/WuRWW022, tvformer}. 
With the development of 5G wireless networks and the artificial intelligence of things (AIoT), 
deploying ViTs on resource-constrained edge devices to enable real-time multimedia applications has become an appealing prospect. 
However, the high computational complexity of ViTs poses a significant challenge to deploy them on edge devices. 
For example, ViT-L/16 \cite{deit}, a typical ViT architecture for computer vision, 
requires over 180 giga FLOPs for inference and takes 56.79 milliseconds on an NVIDIA Jetson TX2 device to classify an image with $224\times 224$ resolution. 
Given that performance and quality-of-service (QoS) are critical for real-time multimedia systems, deploying such latency-greedy ViTs on resource-constrained edge devices is a challenging task.

Early exiting provides a feasible solution for accelerating the inference of neural networks by terminating forward propagation once the prediction from internal classifiers satisfies a certain criterion. 
While early exiting has been extensively studied for CNNs and transformer models in NLP, its application to ViTs remains an open problem. 
The main challenges in developing an efficient early exiting framework for ViTs can be condensed into three key aspects. 
Firstly, directly applying early exiting strategies on ViTs leads to substantial performance degradation. 
However, there has been no systematic investigation into what limits the performance. 
Secondly, minimizing the accuracy drop and further accelerating the inference of early exiting ViTs on edge devices is challenging. 
Lastly, in the training phase, internal classifiers lose considerable information from the final classifier, resulting in poor performance.

Regarding the first challenge,
Kaya \textit{et al.} \cite{SDN} discovered CNNs can reach correct predictions before their final layer, 
and they introduced internal classifiers to mitigate the overthinking problem. 
Sajjad \textit{et al.} \cite{DBLP:journals/csl/SajjadDDN23} examined the impact of dropping layers in transformer models and found that lower layers are more critical for task performance. 
However, their analyses were limited to CNNs or transformer models and did not consider the constraints of early exiting methods in ViTs. 
Concerning the second challenge, a series of studies have introduced exiting criteria to determine when to terminate forward propagation \cite{PCEE, hash} or 
designed advanced backbone networks to balance performance and efficiency \cite{branchynet, ztw}. 
Although Bakhtiarnia \textit{et al.} \cite{ViT-EE} proposed an early exiting framework for ViT by incorporating additional backbone blocks as exiting heads, 
a considerable speed-up gap remains between these methods and the constraints imposed by mobile and edge platforms. 
It is advantageous to design efficient exiting heads for constructing early exiting ViTs with rich feature representations. 
In relation to the third challenge, 
distillation-based \cite{fastbert,self-distill} approaches provide a promising solution to help internal classifiers imitate the final classifiers. 
However, these methods are only available to the same early exiting head architectures.

To remedy these limitations, we initially conduct probing experiments to examine the direct application of early exiting methods in ViTs. 
We discover that the performance of early exiting is constrained by: 
\textit{i)} inadequate feature representations in shallow internal classifiers; 
\textit{ii)} the weak ability to capture target semantic information in deep internal classifiers. 
Building on these insights, we then propose an efficient early exiting framework for general ViTs, termed LGViT, which accelerates inference while maintaining almost the same accuracy.
In LGViT, we incorporate heterogeneous exiting heads, specifically, the local perception head and global aggregation head, 
to generate early exiting ViT networks. 
The local perception head is attached to shallow exiting points to capture local information and learn sufficient feature representations. 
Conversely, the global aggregation head is connected to deep exiting points to extract global information, thereby enhancing the capture of target semantic feature. 
To the best of our knowledge, this is the first work to employ heterogeneous exiting heads for early exiting ViTs. 
Subsequently, we propose a novel two-stage training strategy for early exiting ViTs. 
During the first stage, we utilize an end-to-end method to help the backbone ViT achieve its full potential. 
In the second stage, we froze the parameters of backbone and solely update the exiting heads. 
Self-distillation between exiting heads is employed to minimize information loss. 
Lastly, we perform extensive experiments to validate the superiority of our proposed framework for accelerating ViT inference, 
achieving a good efficiency-accuracy trade-off for three ViT backbones on three datasets. 
For example, as shown in Figure \ref{fig:trade-off}, when ViT serves as the backbone, our method can accelerate the inference by 1.72 $\times$ with only 1.7 \% accuracy drop on the CIFAR-100 dataset. 

Our main contributions are summarized as follows:
\begin{itemize}
    \item We conduct a systematic investigation into the effectiveness of early exiting in ViTs and analyze the issues arising from the vanilla early exiting.
    \item We propose an efficient early exiting framework termed LGViT for general ViTs, incorporating heterogeneous exiting heads, \textit{i.e.}, 
    local perception head and global aggregation head, to achieve an efficiency-accuracy trade-off. 
    \item We develop a novel two-stage training strategy that facilitates learning among multiple heterogeneous exiting heads and significantly minimizes information loss.
    \item We perform extensive experiments on three widely-used datasets and representative ViT backbones, demonstrating the superiority of our proposed framework, 
    which achieves an average speed-up of 1.8 $\times$ with only 2\% accuracy sacrifice. 
\end{itemize}

%% file: sections/related_work.tex
\section{Related works}
\label{sc:related}

\textbf{Efficient ViT. }
Due to their considerable computational cost, ViTs are challenging to deploy on resource-constrained edge devices for real-time inference \cite{DBLP:journals/csur/TayDBM23, shen2023efficient}. 
Recently several studies have proposed lightweight architectures to enhance performance.
For example, Mehta \textit{et al.} \cite{mobilevit} incorporate convolution into transformers, combining the strengths of convolution and attention. 
Maaz \textit{et al.} \cite{DBLP:conf/eccv/MaazSCKZAK22} propose an efficient hybrid architecture and design split depth-wise channel groups encoder to increase the receptive field. 
Furthermore, a series of methods employ traditional model compression techniques to obtain compact ViTs, such as network pruning \cite{DBLP:conf/cvpr/Tang00XGXT22, kwon2022fast, DBLP:conf/nips/ZhenglZYTXRP22}, knowledge distillation \cite{deit, hao2022learning} and low-bit quantization \cite{DBLP:conf/mm/DingQYCLWL22, yuan2022ptq4vit}. 
Hao \textit{et al.} \cite{hao2022learning} utilize patch-level information to help compact student models imitate teacher models. 
Kwon \textit{et al.} \cite{kwon2022fast} propose a post-training pruning framework with structured sparsity methods. 

\noindent
\textbf{Early exiting strategy. }
Early exiting is an effective dynamic inference paradigm that allows confident enough predictions from internal classifiers to exit early. 
Recent research on early exiting can be broadly categorized into two classes: 
\textit{1) Architecture design.} 
Some studies focus on designing advanced backbone networks to balance performance and efficiency. 
For example, Teerapittayanon \textit{et al.} \cite{branchynet} first propose to attach internal classifiers at varying depth in DNNs to accelerate inference. 
Wołczyk \textit{et al.} \cite{ztw} introduce cascade connections to enhance information flow between internal classifiers and aggregate predictions from multiple internal classifiers to improve performance. 
These methods scarcely consider the design of the exiting head architecture and nearly all utilize a fully connected layer following a pooler as the exiting head. 
Bakhtiarnia \textit{et al.} \cite{ViT-EE} propose to insert additional backbone blocks as early exiting branches into ViT.
\textit{2) Training scheme.} 
Another line of work designs training schemes to enhance the performance of internal classifiers. 
Liu \textit{et al.} \cite{fastbert} employ self-distillation to help internal classifiers learn the knowledge from the final classifier. 
Xin \textit{et al.} \cite{BERxiT} introduce an alternating training scheme to alternate between two objectives for odd-numbered and even-numbered iterations. \\

Existing methods for efficient ViT primarily focus on elaborately designing compact ViT structures or applying model compression techniques to compress ViT. 
Our approach adopts sample-level acceleration for inference by dynamically adapting outputs at different paths based on the confidence of each exit's prediction. 
Regarding early exiting strategies, 
the work most related to this paper is CNN-Add-EE and ViT-EE in \cite{ViT-EE}, which use a convolution layer and a transformer encoder as exiting heads, respectively.
However, their performance is unsatisfactory and cannot achieve efficient inference. 
To the best of our knowledge, we first introduce heterogeneous exiting heads to construct early exiting ViTs and achieve an efficiency-accuracy trade-off. 
On top of the aforementioned studies, 
We also propose a novel two-stage training scheme to bride the gap between heterogeneous architectures.

%% file: sections/method.tex
\section{Method}
\label{sc:method}

In this section, we first provide the motivation and an overview of the proposed LGViT framework. 
Then we illustrate the heterogeneous exiting heads and two-stage training strategy. 
Lastly, we depict the exit policy employed during the inference process. 


\subsection{Motivation}
\begin{figure}
  \includegraphics[width=0.38\textwidth]{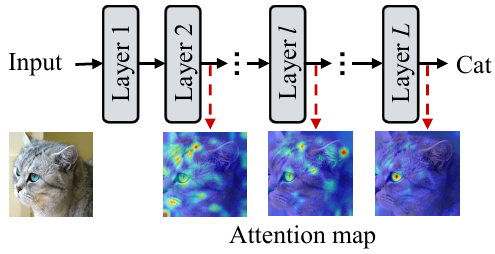}
  \vspace{-8pt}
  \caption{Comparison of attention maps at different exiting positions. 
  The classifiers are omitted. 
  The shallow internal classifiers are difficult to identify object due to inadequate feature capture. 
  The deep internal classifiers do not capture target semantic information compared to the last classifiers. 
  }
  \Description{Motivation.}
  \label{fig:motivation}
  \vspace{-10pt}
\end{figure}

The early exiting method can halt the forward propagation of neural networks prematurely to provide an efficiency-accuracy trade-off, 
which has achieved significant performance improvements for CNNs and transformers in NLP. 
However, naively implementing early exiting on ViT may not yield performance gains for internal classifiers. 
For instance, the performance of the internal classifier on the fifth and tenth layers decreases by 21.8\% and 4.0\%, respectively, 
compared to the original classifier for ViT-B/16 \cite{vit} on CIFAR-100 \cite{cifar}. 
As illustrated in Figure \ref{fig:motivation}, we compare the attention maps at different exiting points for DeiT-B 
(the detailed description of probing experiments is presented in Appendix \ref{sc:investigation}). 
The deep classifiers can extract target semantic features to identify objects. Therefore, 
we obtain the following observations: 
\begin{itemize}
  \item \textbf{\textit{Observation 1}:} Shallow internal classifiers cannot learn sufficient feature representation. 
  \item \textbf{\textit{Observation 2}:} Deep internal classifiers cannot capture target semantic information. 
\end{itemize}

We also discover that if both convolution and self-attention are employed as exiting architectures, positioned on shallow and deep layers respectively, 
the model would gain access to a more comprehensive combination of local and global information compared to the vanilla head architecture. 

\subsection{Overview}

\begin{figure*}
    \includegraphics[width=0.84\textwidth]{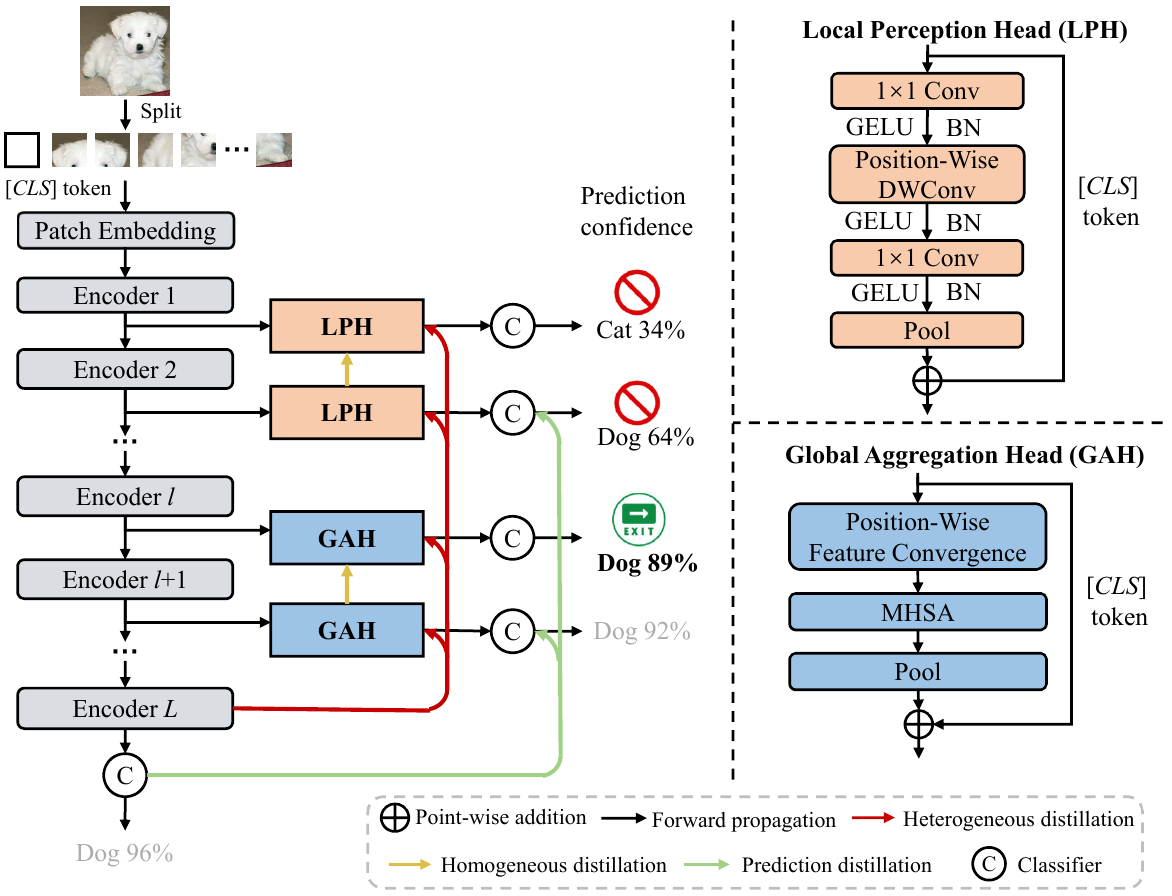}
    \vspace{-6pt}
    \caption{Overview of the proposed early-exiting ViT framework. 
    1) Given a backbone of ViT, we first attach local perception head (LPH) at lower half exiting points and global aggregation head (GAH) at top half of exiting points. 
    2) During the training phase, after an end-to-end training of the backbone, 
    all exiting heads are jointly trained through a novel self-distillation utilizing heterogeneous features, homogeneous features and prediction logits as supervision with the backbone frozen. 
    3) In the inference stage, each input sample dynamically adjusts its exiting path according to the prediction confidence.}
    \Description{An overview image.}
    \label{fig:Overview}
    \vspace{-8pt}
  \end{figure*}

Motivated by the aforementioned observations, we propose an early exiting framework for ViTs that incorporates heterogeneous early exiting architectures.
An overview of the proposed framework is depicted in Figure \ref{fig:Overview}. 
It comprises a ViT backbone, multiple local perception heads and multiple global aggregation heads. 
Initially, a ViT backbone consisting of $L$ encoder blocks is provided. We add $M$ internal classifiers to intermediate blocks of ViT. 
Generally, $M$ is smaller than the total number of backbone layers, as adding internal classifiers after every layer would result in substantial computation costs. 
The position of internal classifiers is independent of the backbone layer numbering. 

We follow a three-step procedure to construct the early exiting ViT framework: 
\begin{itemize}
  \item \textbf{Attaching heterogeneous exiting heads:} Starting from a backbone of ViT, 
  we first select several exiting points along its depth. Then we place local perception heads and global aggregation heads at corresponding exiting points according to their positions. 
  \item \textbf{Two-Stage training:} We train the whole early exiting ViT using a novel two-stage training strategy including end-to-end training and self-distillation with the backbone frozen.  
  This can facilitate the integration of global and local information for performance improvement. 
  \item \textbf{Dynamic inference:} When the trained model is deployed for inference, 
  each input sample can dynamically determine its output at varying depths based on the confidence of each exiting head prediction. 
\end{itemize}

\subsection{Attaching Heterogeneous Exiting Heads}
We first introduce the placement of exiting heads, followed by a detailed description of the heterogeneous exiting heads. 
In this work, exiting points, where exiting heads are positioned, are determined according to an approximately equidistant computational distribution, 
\textit{i.e.}, the multiply-accumulate operations (MACs) of intermediate blocks between two adjacent points remain consistent. 
For the sake of simplicity, exiting points are constrained to be placed at the output of individual encoder blocks.
We attach local perception heads, based on convolution, to the lower half of exiting points to enhance local information exploration. 
Global aggregation heads, based on self-attention, are integrated into the upper half of points to augment global information acquisition. 

\textbf{Local perception head.} 
As analyzed in Section \ref{sc:investigation}, directly applying early exiting in ViT leads to severe performance degradation for shallow internal classifiers. 
To mitigate the issue, we introduce a local perception head (LPH) for early exiting framework, which elegantly incorporates convolution into ViT to enhance feature representation learning. 
It can achieve efficient local information exploration and effective feature integration extracted from the original backbone.   
As illustrated in the upper right of Figure \ref{fig:Overview}, 
the proposed LPH first employs a $1\times 1$ convolution layer to expand dimensions. 
Subsequently, the expanded features are passed to a position-wise depth-wise convolution (PDConv) with $k\times k$ kernel size that depends on the exiting positions $m$.
In order to reduce computation overhead, we employ smaller kernel size convolutions for the deeper exiting points. 
We employ a decreasing linear mapping function $f(\cdot)$ to determine the kernel size of PDConv, \textit{i.e.}, $k=f(m), m\leq L/2$. 
For instance, the expanded features at the $m$-th exiting position are passed to $k\times k$ depth-wise convolution. 
Note that $k=0$ means that the expanded features will bypass PDConv and proceed directly to the subsequent part. 
Thus, the PDConv can be formulated as: 
\begin{equation}
  \text{PDConv}(\mathbf{X}, m)=\begin{cases}
    \text{DWConv}_{k\times k}(\mathbf{X}), \quad & f(m) > 0 \\
    \mathbf{X}, \quad & f(m) = 0
  \end{cases},
\end{equation}
where $\text{DWConv}_{k\times k}$ denotes the depth-wise convolution with kernel of size $k\times k$. 
Then the features are projected back into the original patch space using a $1\times 1$ convolution and then passed to an average pooling layer. 
Considering that the $[CLS]$ token contains djominant feature representations, 
it is added to the pooled output to facilitate the fusion of global information from the original backbone and local information from the convolution.
Concretely, 
given a ViT encoder output $\mathbf{X}_{en} \in \mathbb{R}^{N\times D} $, 
where $N$ represents the number of patches and $D$ denotes the hidden dimension, 
when the position of exiting points is lower than $L/2$, the output of the proposed exiting head is given by: 
\begin{equation}
  \begin{aligned}
    \text{LPH}(\mathbf{X}_{en}, m)&=\text{Pool}(\text{Conv}_{1\times 1}(\mathcal{G}(\mathbf{X}_{en}, m) ) )+\mathbf{X}_{CLS}, \\
    \mathcal{G}(\mathbf{X}_{en}, m)&=\text{PDConv}(\text{Conv}_{1\times 1}(\mathbf{X}_{en}), m),
  \end{aligned}
\end{equation}
where $\mathbf{X}_{CLS}$ represents $[CLS]$ token and the activation layer is omitted. 
Gaussian error linear unit (GELU) and batch normalization (BN) are employed after each convolution.
The output of LPH is finally passed to the internal classifier. 
By introducing LPH as the exiting head, shallow internal classifiers can learn adequate feature representation and capture local information, thereby enhancing performance in vision tasks. 

\textbf{Global aggregation head.} 
Based on the discussion in Section \ref{sc:investigation}, the direct application of early-exit methods to ViT hinders the semantic information capture in deep internal classifiers. 
We propose a global aggregation head (GAH) and incorporate it at deep exiting points, as illustrated in the lower right of Figure \ref{fig:Overview}. 
The proposed GAH integrates features from locally adjacent tokens and then compute self-attention for each subset to facilitate target semantic information exploitation. 
In GAH,  
we first employ a position-wise feature convergence (PFC) block to aggregate features from the exiting point. 
In the PFC block, input features $\mathbf{X}\in \mathbb{R}^{N\times D}$ are reshaped to $D\times H\times W $ dimensions and down-sampled with a size of $s\times s$ window.
The sampled features $\mathbf{X}_{sample}\in \mathbb{R}^{D\times \frac{H}{s} \times \frac{W}{s}}$ are restored to original dimension format $\frac{N}{s^2}\times D$. 
The proposed PFC block reshapes the input features to patch format and down-samples them with an $s\times s$ window. 
The sampled features are then restored to original format. 
To avoid introducing additional learnable parameters, we employ an average pool with $s$ stride as the implementation of PFC. 
Analogous to PDConv, the swindow size $s$ of PFC also depends on the exiting position $m$. 
Deeper exiting points utilize larger window sizes significantly reducing the computational cost. 
We employ an increasing linear mapping function $g(\cdot)$ to determine the window size of PFC, \textit{i.e.}, $s=g(m), L/2 < m \leq L$. 
For example, the input features are passed to sub-sample with a size of $g(m)\times g(m)$ window at the $m$-th exiting point. 
Note that the minimum window size is generally set to $2$. 
Consequently, the PFC can be expressed as:
\begin{equation}
  \text{PFC}(\mathbf{X}_{en}, m)=\text{Pool}_{g(m)}(\mathbf{X}_{en}), 
\end{equation}
where $\text{Pool}_{g(m)}$ represents an average pool with $g(m)$ stride. The reshaping and recover operations of features are omitted in the equation. 
PFC not only reduces the computational redundancy but also helps focus on target patches compared to the original MHSA. 
Then the integrated features are passed through multi-head self-attention (MHSA) and a pool layer. 
The $[CLS]$ token is also added to the pooled features. 
Thus, when the position of exiting points is deeper than $L/2$, the proposed GAH can be formulated as:
\begin{equation}
  \begin{aligned}
    \text{GAH}(\mathbf{X}_{en}, m)&=\text{Pool}(\text{MHSA}(\text{PFC}(\mathbf{X}_{en}, m)))+\mathbf{X}_{CLS}, \\
    \text{MHSA}(\mathbf{X})&=\text{softmax}\left(\frac{XW_{Q}(XW_{K})^T}{\sqrt{d}}\right)XW_{V},
  \end{aligned}
\end{equation}
where the input $\mathbf{X}$ is linearly transformed into query, key and value vectors using transformation matrices $W_{Q}$, $W_{K}$ and $W_{V}$; 
$d$ is the vector dimension. 
By employing GAH as the exiting head, deep internal classifiers can reduce spatial redundancy of self-attention and capture more target semantic information. 

\textbf{Complexity analysis.} 
To thoroughly understand the computational bottleneck of heterogeneous exiting heads, 
we compare our proposed LPH + GAH with standard convolution + MHSA by analyzing their floating-point operations (MACs). 
Given an input feature of size $N\times D$, the FLOPs of standard $k\times k$ convolution are:
\begin{equation}
  \mathcal{O} (\text{Conv}_{k\times k})=ND^2k^2. 
\end{equation}
The FLOPs of an MHSA module can be calculated as:
\begin{equation}
  \mathcal{O} (\text{MHSA})=2ND(D+D)+N^2(D+D)=4ND^2+2N^2D, 
\end{equation}
where the activation function is omitted. 
For a fair comparison, we select the same kernel size $k$ in LPH. The FLOPs of the proposed two exiting head are as follows:
\begin{equation}
  \begin{aligned}
    \mathcal{O}(\text{LPH})&=2ND^2+NDk^2, \\
    \mathcal{O}(\text{GAH})&=4ND^2/s^2+2N^2D/s^4.
  \end{aligned}
\end{equation}
We observe that the computational complexity of LPH and GAH is lower than that of standard convolution and MHSA, respectively. 
\begin{equation}
  \begin{aligned}
    \frac{\mathcal{O}(\text{LPH})}{\mathcal{O} (\text{Conv}_{k\times k})}&=(2D+k^2)/Dk^2<1, \\
    \frac{\mathcal{O}(\text{GAH})}{\mathcal{O} (\text{MHSA})}&=\frac{2D+N/s^2}{2D+N}<1.
  \end{aligned}
\end{equation}
Therefore, compared to standard convolution and MHSA, our proposed LPH and GAH heads are more friendly to computational cost and convenient to implement on hardware platforms.

\subsection{Two-Stage Training Strategy}
To tackle the performance degradation issue caused by early exiting, 
we propose a novel two-stage training strategy to transfer knowledge from deeper classifiers to shallow classifiers. 
The process of two-stage training strategy is presented as follows. 
\begin{itemize}
  \item In the first stage, we train the backbone ViT and update the parameters of the backbone and the final classifier in an alternating strategy. 
  As a result, the interference of multiple internal classifiers can be minimized, enabling the final classifier to reach its full potential. 
  Similar to general training, we utilize the cross entropy function as the training loss. 
  \item During the second stage, the backbone and final classifier are kept frozen. Only the parameters of exiting heads and internal classifiers can be updated. 
  We introduce self-distillation to facilitate the imitation of all exiting heads from the last classifier as illustrated in the left of Figure \ref{fig:Overview}. 
\end{itemize}

The overall distillation loss comprises the heterogeneous distillation, homogeneous distillation and prediction distillation loss.

\textbf{Heterogeneous distillation. }
Considering the usage of heterogeneous exiting head, the gap between different types of heads is substantial. 
Directly utilizing the features from the last layer as supervision for internal classifiers can lead to information loss. 
To tackle this issue, we propose heterogeneous distillation to facilitate learning the knowledge from heterogeneous architectures. 
To reduce the conflict between multiple losses, we only employ the feature of the last layer as the reference feature for the first and last exiting heads of LPH and GAH. 
The inductive bias and local visual representations captured by LPH can be elegantly integrated with global information extracted from self-attention. 
Considering the different shapes of feature maps between exiting heads and the final block, we employ an aligning module to match the dimensions. 
The module consists of a depth-wise convolution, GELU and BN activation functions. 
The feature map of the last ViT layer $\mathbf{F}_L \in \mathbb{R}^{N\times D}$ is first reshaped to $D\times \sqrt{N}\times \sqrt{N}$ dimensions. 
The reshaped feature map is passed to the module to reduce dimensions, and then restored to original dimension format $N'\times D$. 
The loss function of heterogeneous can be formulated as: 
\begin{equation}
  \mathcal{L}_{hete}=\frac{1}{4}\sum_{m \in \mathcal{M} }\mathcal{L}_{KL}(\mathbf{F}_{m}, \text{Align}(\mathbf{F}_L)), 
\end{equation}
where $\mathcal{M}= \{1,M/2,M/2+1,M\}$ and $\mathcal{L}_{KL}$ is the Kullback-Leibler divergence function. 

\textbf{Homogeneous distillation. }
We propose homogeneous distillation between exiting heads with the same architectures to further improve performance. 
In each type of exiting heads, we employ the final heads as the teacher assistant to help the preceding homogeneous heads learn hint knowledge. 
For example, 
in all LPHs, the features of final LPH (\textit{i.e.} at $M/2$-th exiting point) is utilized as reference feature for the preceding LPH. 
Given the feature maps from the first to $m$-th exiting heads $\mathbf{F}_m, (1\leq m\leq M/2)$, 
the loss function of homogeneous distillation between LPHs is:
\begin{equation}
  \mathcal{L}_{homo}^{LPH}=\frac{1}{M/2-1}\sum_{m = 1}^{M/2-1} \mathcal{L}_{MSE}(\mathbf{F}_m, \mathbf{F}_{M/2}), 
\end{equation}
where $\mathcal{L}_{MSE}$ is the mean squared error function. 
Since the feature maps of GAH have different shapes, we apply dot-product operations between feature maps. 
Given a feature map of GAH $\mathbf{F}_m\in \mathbb{R}^{N/g^2(m) \times D}$ at the $m$-th exiting point, 
the shape can be transformed to $D\times D$ by computing $\mathbf{F}_m^T\mathbf{F}_m$. 
The loss function of homogeneous distillation between GAHs can be expressed as: 
\begin{equation}
  \mathcal{L}_{homo}^{GAH}=\frac{1}{M/2-1}\sum_{m = M/2+1}^{M-1} \mathcal{L}_{MSE}(\mathbf{F}_m^T\mathbf{F}_m, \mathbf{F}_{M}^T\mathbf{F}_{M}). 
\end{equation}
Therefore, the overall loss function of homogeneous distillation can be expressed as: 
\begin{equation}
  \mathcal{L}_{homo}=\mathcal{L}_{homo}^{LPH}+\mathcal{L}_{homo}^{GAH}. 
\end{equation}

\textbf{Prediction distillation. }
In order to further improve the performance of internal classifiers, 
we utilize the final classifier as the reference label of $M/2$-th and $M$-th exiting points where last LPH and GAH are located, respectively. 
Given an input sample associated with label $y$, and assuming that the predictions at the $M/2$-th and $M$-th exiting points are $\hat{y}_{M/2}$ and $\hat{y}_{M}$, respectively, 
the loss function of prediction distillation can be formulated as: 
\begin{equation}
  \mathcal{L}_{pred}=\mathcal{L}_{KD}(\hat{y}_{M/2}, \hat{y}_{L}, y)+\mathcal{L}_{KD}(\hat{y}_{M}, \hat{y}_{L}, y),
\end{equation}
where $\mathcal{L}_{KD}$ is the loss function of vanilla knowledge distillation: 
\begin{equation}
  \mathcal{L}_{KD}(\hat{y}^s, \hat{y}^t, y) = (1-\gamma)\mathcal{L}_{CE}(\hat{y}^s, y) + \gamma \mathcal{L}_{KL}(\hat{y}^s/T, \hat{y}^t/T).  
\end{equation}
Here, $\mathcal{L}_{CE}$ is the cross-entropy function, 
$T$ is a temperature value to control the smoothness of logits, 
and $\gamma$ is a balancing hyperparameter. 

Hence, the overall loss function of our proposed method is:
\begin{equation}
  \mathcal{L} = \alpha \mathcal{L}_{hete} + \beta \mathcal{L}_{homo} + \mathcal{L}_{pred}, 
\end{equation}
where $\alpha$ and $\beta$ are hyperparameters.

\subsection{Dynamic Inference}
In this section, we first introduce the exiting metric and then depict the process of early exiting ViT inference. 
We employ a standard confidence metric following \cite{ztw} as the exiting metric, which represents the probability of the most confident classification class. 
The prediction confidence $c_m$ at the $m$-th exiting position is:
\begin{equation}
  c_m(p^m)= \mathop{max}\limits_{C }p^m,
\end{equation}
where $p^m$ is the prediction distribution at $m$-th exiting position and $C$ is the classification label set. 
During the inference process, input samples go through exits sequentially. 
Each sample dynamically adjusts its exiting path according to the exiting metric. 
If the classification confidence of a sample at the $m$-th exiting point exceeds a predefined threshold $\tau$, the forward propagation of ViT will be terminated, 
and the prediction at the $m$-th exiting point will be output. 
The threshold $\tau$ can be adjusted according to computation cost and hardware resources to achieve an efficiency-accuracy trade-off. 
A low threshold may lead to a significant speed-up at the cost of a possible drop in accuracy.
If the exiting condition is never reached, the ViT will revert to the standard inference process.

\begin{table*}
  \renewcommand{\arraystretch}{0.9}
  \centering
  \setlength{\extrarowheight}{0pt}
  \addtolength{\extrarowheight}{\aboverulesep}
  \addtolength{\extrarowheight}{\belowrulesep}
  \setlength{\aboverulesep}{0pt}
  \setlength{\belowrulesep}{0pt}
  \caption{Performance of different methods on three datasets for different ViT backbones. "Acc." represents the Top-1 classification accuracy. "\#Params." represents the number of model parameters.}
  \label{tb:performance}
  \vspace{-7pt}
  \begin{tabular}{c|c|ccc|ccc|ccc} 
  \toprule
  \multirow{2}{*}{\textbf{Methods}}               & \multirow{2}{*}{\textbf{\#Params.}} & \multicolumn{3}{c|}{\textbf{Results: CIFAR-100}}                             & \multicolumn{3}{c|}{\textbf{Results: Food-101}}                              & \multicolumn{3}{c}{\textbf{Results: ImageNet-1K}}                             \\
                                                  &                                     & \textbf{Acc.~}   & \textbf{MACs~$\downarrow$} & \textbf{Speed-up~$\uparrow$} & \textbf{Acc.}    & \textbf{MACs~$\downarrow$} & \textbf{Speed-up~$\uparrow$} & \textbf{Acc.}    & \textbf{MACs~$\downarrow$} & \textbf{Speed-up~$\uparrow$}  \\ 
  \hline
  \multicolumn{11}{c}{\textbf{ViT}}                                                                                                                                                                                                                                                                                                   \\ 
  \hline
  ViT-B/16                                        & 86 M                                & 90.8 \%          & 16.93 G                    & 1.00~$\times$                & 89.6 \%          & 16.93 G                    & 1.00~$\times$                & 81.8 \%          & 16.93 G                    & 1.00~$\times$                 \\ 
  \hline
  SDN                                             & 94 M                                & 86.5 \%          & 10.16 G                    & 1.64~$\times$                & 88.5 \%          & 8.67 G                     & 1.95~$\times$                & 79.5 \%          & 10.95 G                    & 1.55~$\times$                 \\
  PABEE                                           & 94 M                                & 85.1 \%          & 11.48 G                    & 1.52~$\times$                & 86.7 \%          & 10.93 G                    & 1.81~$\times$                & 78.6 \%          & 12.41 G                    & 1.36~$\times$                 \\
  BERxiT                                          & 94 M                                & 87.1 \%          & 10.27 G                    & 1.65~$\times$                & 88.3 \%          & 8.56 G                     & 1.98~$\times$                & 79.9 \%          & 11.75 G                    & 1.44~$\times$                 \\
  ViT-EE                                          & 94 M                                & 87.5 \%          & 11.65 G                    & 1.65~$\times$                & 88.2 \%          & 10.42 G                    & 1.91~$\times$                & 79.6 \%          & 13.66 G                    & 1.38~$\times$                 \\
  PCEE                                            & 94 M                                & 86.1 \%          & 10.90 G                    & 1.55~$\times$                & 88.1 \%          & 9.50 G                     & 1.81~$\times$                & 80.0 \%          & 12.36 G                    & 1.37~$\times$                 \\
  \rowcolor[rgb]{0.949,0.949,0.949} \textbf{Ours} & 101 M                               & \textbf{88.5} \% & \textbf{9.76 G}            & \textbf{1.87}~$\times$       & \textbf{88.6} \% & \textbf{7.63 G}            & \textbf{2.36}~$\times$       & \textbf{80.3}~\% & \textbf{10.65 G}           & \textbf{1.70}~$\times$        \\ 
  \hline
  \multicolumn{11}{c}{\textbf{DeiT}}                                                                                                                                                                                                                                                                                                  \\ 
  \hline
  DeiT-B\alambic\xspace                                  & 86 M                                & 91.3 \%          & 16.93 G                    & 1.00~$\times$                & 90.3 \%          & 16.93 G                    & 1.00~$\times$                & 83.4 \%          & 16.93 G                    & 1.00~$\times$                 \\ 
  \hline
  SDN                                             & 94 M                                & 87.4 \%          & 9.65 G                     & 1.75~$\times$                & 88.5 \%          & 8.62 G                     & 1.97~$\times$                & 77.5 \%          & 11.30 G                    & 1.50~$\times$                 \\
  PABEE                                           & 94 M                                & 86.4 \%          & 11.43 G                    & 1.48~$\times$                & 88.6 \%          & 11.00 G                    & 1.54~$\times$                & 78.5 \%          & 12.40 G                    & 1.36~$\times$                 \\
  BERxiT                                          & 94 M                                & 88.3 \%          & 10.48 G                    & 1.61~$\times$                & 88.8 \%          & 9.16 G                     & 1.85~$\times$                & 79.1 \%          & 11.12 G                    & 1.53~$\times$                 \\
  ViT-EE                                          & 93 M                                & 88.3 \%          & 11.07 G                    & 1.75~$\times$                & 88.8 \%          & 10.26 G                    & 1.91~$\times$                & 80.5 \%          & 13.28 G                    & 1.45~$\times$                 \\
  PCEE                                            & 94 M                                & 87.5 \%          & 10.49 G                    & 1.61~$\times$                & 88.6 \%          & 9.59 G                     & 1.76~$\times$                & 80.4 \%          & 11.87 G                    & 1.43~$\times$                 \\
  \rowcolor[rgb]{0.949,0.949,0.949} \textbf{Ours} & 102 M                               & \textbf{88.9} \% & \textbf{9.54 G}            & \textbf{1.91}~$\times$       & \textbf{89.5} \% & \textbf{8.53 G}            & \textbf{2.12}~$\times$       & \textbf{81.7}~\% & \textbf{10.90 G}           & \textbf{1.67}~$\times$        \\ 
  \hline
  \multicolumn{11}{c}{\textbf{Swin}}                                                                                                                                                                                                                                                                                                  \\ 
  \hline
  Swin-B                                          & 87 M                                & 92.6 \%          & 15.13 G                    & 1.00~$\times$                & 93.3 \%          & 15.13 G                    & 1.00~$\times$                & 83.5 \%          & 15.40 G                    & 1.00~$\times$                 \\ 
  \hline
  SDN                                             & 88 M                                & 88.3 \%          & 9.41 G                     & 1.72~$\times$                & 90.2 \%          & 7.64 G                     & 2.17~$\times$                & 78.7 \%          & 10.91 G                    & 1.45~$\times$                 \\
  PABEE                                           & 88 M                                & 83.8 \%          & 9.46 G                     & 1.72~$\times$                & 88.8 \%          & 8.17 G                     & 2.01~$\times$                & 79.0 \%          & 13.19 G                    & 1.18~$\times$                 \\
  BERxiT                                          & 88 M                                & 88.4 \%          & 9.61 G                     & 1.68~$\times$                & 90.2 \%          & 7.68 G                     & 2.16~$\times$                & 80.2 \%          & 10.35 G                    & 1.54~$\times$                 \\
  ViT-EE                                          & 91 M                                & 88.1 \%          & 9.71 G                     & 1.82~$\times$                & 90.6 \%          & 8.92 G                     & 2.03~$\times$                & 82.1 \%          & 11.20 G                    & 1.52~$\times$                 \\
  PCEE                                            & 88 M                                & 88.1 \%          & 10.68 G                    & 1.50~$\times$                & 90.3 \%          & 9.11 G                     & 1.79~$\times$                & 79.8 \%          & 11.51 G                    & 1.37~$\times$                 \\
  \rowcolor[rgb]{0.949,0.949,0.949} \textbf{Ours} & 97 M                                & \textbf{90.7} \% & \textbf{8.84 G}            & \textbf{1.94}~$\times$       & \textbf{91.9} \% & \textbf{7.05 G}            & \textbf{2.50}~$\times$       & \textbf{82.7} \% & \textbf{9.98 G}            & \textbf{1.69}~$\times$        \\
  \bottomrule
  \end{tabular}
  \vspace{-7pt}
  \end{table*}

%% file: sections/experiment.tex
\section{Experiment}
\label{sc:ex}

In this section, we first introduce some implementation details and experimental settings. 
Then, we present the results of performance evaluations on three vision datasets and three popular ViT backbones. 
Finally, we conduct extensive ablation experiments to demonstrate the superiority of our methods. 

\subsection{Experimental Setup}

\textbf{Datasets.}
We evaluate our proposed method on three public vision datasets: 
CIFAR-100\cite{cifar100}, Food-101\cite{food101}, and ImageNet-1K\cite{imagenet}. 
The CIFAR-100 dataset contains 50K training images and 10K testing images, uniformly categorized into 100 classes. 
The Food-101 dataset consists of 101 food categories, with 750 training and 250 test images per category, making a total of 101K images. 
The ImageNet-1K dataset spans 1000 object classes and contains 1,281,167 training images, 50K validation images and 100K test images. 
We augment the training data with random crops, random horizontal flips and normalization, while the testing data is augmented with center crops and normalization.

\noindent
\textbf{Backbones.}
The proposed framework can be applied to a range of early-exit ViTs. Without loss of generality, 
we conduct experiments with three well-known ViT backbones, namely, ViT\cite{vit}, DeiT\cite{deit}, and Swin\cite{swin}. 
ViT is the first pure transformer structure for computer vision tasks and utilize a $[CLS]$ token to serve as the image representation for classification tasks. 
DeiT adds an additional distillation token to learn hard labels from the teacher model compared to ViT. 
Swin builds hierarchical feature maps by merging image patches in deeper layers.
The $[CLS]$ token in LPH and GAH is replaced by the encoder output because it contains no $[CLS]$ token.

\noindent
\textbf{Baselines.}
We compare our dynamic early exiting methods with several representative early exiting methods. 
Considering most methods designed for CNNs and transformers in NLP, we transfer these methods to ViT for fair comparison. 
\begin{itemize}
    \item \textbf{SDN} \cite{SDN} utilizes a weighted training strategy and employs a confidence-based criterion to decide whether to exit. 
    \item \textbf{PABEE} \cite{PABEE} employs a patience metric based on the consistency of classification decisions over several internal classifiers to make early exiting decisions. 
    \item \textbf{BERxiT} \cite{BERxiT} introduces an alternating training strategy to train the whole model. 
    \item \textbf{ViT-EE} \cite{ViT-EE} utilizes a ViT encoder layer as its exiting head with the confidence criterion to decide whether to exit. 
    \item \textbf{PCEE} \cite{PCEE} utilizes a patience\&confidence criterion according to the enough number of confident predictions from consecutive internal classifiers. 
    
  \end{itemize}
Unless otherwise specified, the default exit architecture of baselines is a single fully connected layer that follows a pooler.

\noindent
\textbf{Evaluation metrics.}
Considering the trade-off between performance and efficiency, we employ Top-1 classification accuracy and speed-up ratio as the performance and efficiency metric, respectively. 
Since the measurement of runtime might not be stable, we follow \cite{deebert} to calculate the speed-up ratio by comparing the actually executed layers in forward propagation and the complete layers. 
For an $L$-layer ViT, the speed-up ratio is defined as:
\begin{equation}
  \text{Speed-up}=\frac{\sum_{i = 1}^{L}L\times m^i  }{\sum_{i = 1}^{L}i\times m^i}, 
\end{equation}
where $m^i$ is the number of samples that exit at the $i$-th layer of ViT. 
For clarity, we utilize the average multiply-accumulate operations (MACs) performed across the entire test dataset as a metric to assess the computational cost associated with a given model.

\noindent
\textbf{Implementation details.}
Our framework and all the compared methods are implemented using the Huggingface transformer library \cite{huggingface} for fair comparison. 
Most hyperparameters, such as learning rate, optimizer, and dropout probabilities are kept unchanged from the original backbones for fair comparison. 
We list different hyperparameters in the Appendix due to limited space. 
Each network is fine-tuned by 100 epochs on 3 NVIDIA 3090 GPUs, with a batch size of 64. 
There is no early stopping and the checkpoint after full fine-tuning is chosen. 

\subsection{Performance Evaluation}
\label{sc:peformance}

We conduct extensive experiments to compare our methods with the state-of-the-art methods for three ViT backbones on three vision datasets. 
Then we present the performance and efficient trade-off compared with other baselines.

\textbf{Comparison with the state-of-the-art.}
We compare the performance between our methods and baselines on CIFAR-100, Food-101 and Tiny ImageNet datasets when different backbones are adopted, including ViT, DeiT, and Swin.
The results are shown in Table \ref{tb:performance}. 
The original models for different backbones are ViT-L/16, DeiT-B and Swin-B respectively. 
We can find that our method can achieve approximately a 1.8 $\times$ speed-up ratio with only a 2\% accuracy drop compared to the original models on most datasets, 
which significantly outperforms other baselines. 

\textbf{Performance and efficiency trade-off. }
To further verify the robustness and efficiency of our method, we visualize the performance and efficiency trade-off curves in Figure \ref{fig:trade-off} on CIFAR-100 test set. 
The original backbone model is ViT-B/16. We compare five competitive baselines in the dynamic inference scenario. 
We can see that the performance of most early exiting methods drops dramatically when the speed-up ratio increases. 
This also reflects directly applying early exiting methods in ViT leads to unstable performance which cannot meet the requirements of real-time systems. 
However, our method is more robust to the variance of speed-up. 
If we set the almost same speed-up ratio, the accuracy drop of our method is 3.6 \% lower than SDN method. 
When the accuracy is approximate to other baselines, our method can achieve faster speed-up. 
Moreover, our method can dynamically adjust the speed-up ratio without retraining, which is more feasible and friendly. 

\begin{table}
  \renewcommand\arraystretch{0.9}
  \centering
  \caption{Ablation study results of main components. The "$\checkmark$" mark indicating that we adopt the corresponding component. 
  The opposite of LPH and GAH is fully connected layer with a pooler. The opposite of two-stage training is vanilla training. }
  \label{tb:ablation}
  \vspace{-6pt}
  \begin{tabular}{ccccc} 
    \toprule
    \textbf{LPH} & \textbf{GAH} & \textbf{Two-Stage training} & \textbf{Acc.} & \textbf{Speed-up}  \\ 
    \hline
    $\times$    &  $\times$    &      $\times$              &   87.3 \%           &   1.56 ~$\times$                \\
    $\times$     & $\times$     &     $\checkmark$                 &   88.2 \%           &   1.57~$\times$                 \\
    $\times$     &  $\checkmark$     &   $\checkmark$              &    88.3 \%          &   1.71~$\times$                  \\
    $\checkmark$    &  $\checkmark$  &     $\checkmark$             &   \textbf{88.5} \%           &   \textbf{1.87}~$\times$                 \\
    \bottomrule
    \end{tabular}
    \vspace{-6pt}
  \end{table}

  \begin{table}
    \renewcommand\arraystretch{0.9}
    \centering
    \setlength{\extrarowheight}{0pt}
    \addtolength{\extrarowheight}{\aboverulesep}
    \addtolength{\extrarowheight}{\belowrulesep}
    \setlength{\aboverulesep}{0pt}
    \setlength{\belowrulesep}{0pt}
    \caption{Comparison of different exiting head architectures on CIFAR-100. 
    MLP, Conv and Attention refers to utilizing a fully-connected layer, a $3\times 3$ standard convolution layer, and a MHSA block respectively. }
    \label{tb:head}
    \vspace{-6pt}
    \begin{tabular}{ccccc} 
      \toprule
      \textbf{Exiting head}                           & \textbf{\#Params.} & \textbf{MACs} & \textbf{Acc.} & \textbf{Speed-up}  \\ 
      \hline
      MLP \cite{deebert}                                            & 91 M         & 10.78 G              &   88.2 \%               &  1.57~$\times$                    \\
      Conv \cite{ztw}                                           &   129 M         &  13.86 G             & 87.5 \%                  &  1.72~$\times$                  \\
      Attention \cite{ViT-EE}                                      & 105 M          & 12.76 G              &  88.0 \%                & 1.58~$\times$                     \\
      \rowcolor[rgb]{0.949,0.949,0.949} \textbf{Ours} &    101 M        & \textbf{9.76 G}              &  \textbf{88.5} \%            &  \textbf{1.87}~$\times$                    \\
      \bottomrule
      \end{tabular}
      \vspace{-6pt}
    \end{table}

\subsection{Ablation Study}
To fully understand the impact of each part of the proposed framework, we conduct ablation study, 
where all experiments are evaluated on CIFAR-100 and utilize ViT as the backbone. 
We first study the effectiveness of the main components, and then analyze the impact of different exiting head architectures and training schemes. 
Finally, we verify the robustness of our methods for different numbers of predefined exiting heads. 

\textbf{Ablation of main components.} 
We design experiments to verify the effectiveness of the proposed LPH, GAH and two-stage training scheme. 
Table \ref{tb:ablation} presents the accuracy and speed-up ratio utilizing different components. 
The results show that for early exiting in ViT, 
heterogeneous exiting heads and two-stage training scheme are both significant. 
Specifically, we can observe that the two-stage training scheme can significantly improve accuracy. 
Moreover, when combined with LPH and GAH, the inference efficiency and accuracy can be further improved.

\textbf{The architecture of exiting heads.} 
In order to verify the effectiveness of the proposed heterogeneous exiting heads, we compare other competitive architectures using the same two-stage training scheme, as shown in Table \ref{tb:head}. 
We can observe that the proposed heterogeneous exiting heads are crucial to achieve a speed-accuracy trade-off. 
Although utilizing MLP as exiting heads can gain approximate accuracy to ours, the speed-up ratio is low. 
The attention method can achieve a close trade-off between accuracy and speed but with high storage and computation requirements.

\begin{table}
  \renewcommand\arraystretch{0.9}
  \centering
  \setlength{\extrarowheight}{0pt}
  \addtolength{\extrarowheight}{\aboverulesep}
  \addtolength{\extrarowheight}{\belowrulesep}
  \setlength{\aboverulesep}{0pt}
  \setlength{\belowrulesep}{0pt}
  \caption{Comparison of different training schemes with the proposed exiting heads on CIFAR-100. }
  \label{tb:train}
  \vspace{-6pt}
  \begin{tabular}{ccc} 
  \toprule
  \textbf{Training scheme}                        & \textbf{Accuracy} & \textbf{Speed-up}  \\ 
  \hline
  Normal \cite{branchynet}                                         & 86.9 \%              & 1.82 $\times$                   \\
  Weighted \cite{SDN}                                       & 87.4 \%           & 1.80 $\times$                   \\
  Distillation \cite{fastbert}                                   & 86.9 \%           & 1.83 $\times$                   \\
  Alternating \cite{BERxiT}                                    & 87.9 \%           & 1.84 $\times$                  \\
  \rowcolor[rgb]{0.949,0.949,0.949} \textbf{Ours} & \textbf{88.5} \%           & \textbf{1.87} $\times$                   \\
  \bottomrule
  \end{tabular}
  \vspace{-6pt}
  \end{table}

  \begin{figure}
    \begin{center}
      \includegraphics[width=0.4\textwidth, height=0.32\textwidth]{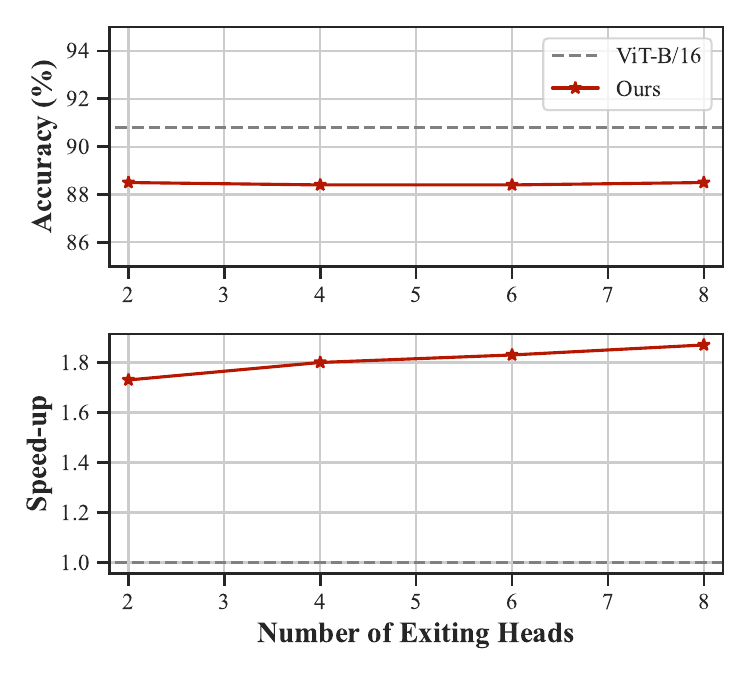}
    \end{center}
    \vspace{-8pt}
    \caption{Results of accuracy and speed-up for different head number settings. The exiting heads are placed across all layers uniformly. }
    \Description{.}
    \label{fig:head_num}
    \vspace{-15pt}
  \end{figure}

\textbf{Training schemes.} 
We compare our methods with four representative training schemes in early exiting methods, and they all utilize the proposed heterogeneous exiting heads. 
The accuracy and speed-up are shown in Table \ref{tb:train}. 
Our method achieves the highest classification accuracy and inference speed-up ratio, which significantly outperforms other training schemes.

  \textbf{The number of exiting heads.} 
  We analyze the influence on accuracy and speed-up when changing the number of predefined exiting heads, as shown in Figure \ref{fig:head_num}. 
  As the number of heads increases, the accuracy remains essentially consistent with only approximate 2 \% accuracy drop, 
  which shows that our method is robust to the number of heads and can tackle the overthinking problem \cite{SDN}. 
  Moreover, we can observe that the speed-up ratio can enhance with the increasing number of heads.

%% file: sections/conclusion.tex
\section{Conclusion}
\label{sc:conclusion}

In this paper, we point out that naively applying early exiting in ViTs results in performance bottleneck due to insufficient feature representations in shallow internal classifiers and limited ability to capture target semantic information in deep internal classifiers. 
Based on this analysis, we propose an early exiting framework for general ViTs which combines heterogeneous exiting heads to enhance feature exploration. 
We also develop a novel two-stage training strategy to reduce information loss between heterogeneous exiting heads. 
We conduct extensive experiments for three ViT backbones on three vision datasets, 
demonstrating that our methods outperform other competitive counterparts. 
The limitation of our methods is to manually choose the exiting position and optimal exiting path. 
In the future, we intend to utilize Bayesian optimization to automatically perform the optimal exiting decision. 

%% file: sections/appendix.tex
\clearpage

\section{Appendix}

\begin{figure}[t]
  \begin{center}
    \includegraphics[width=0.475\textwidth]{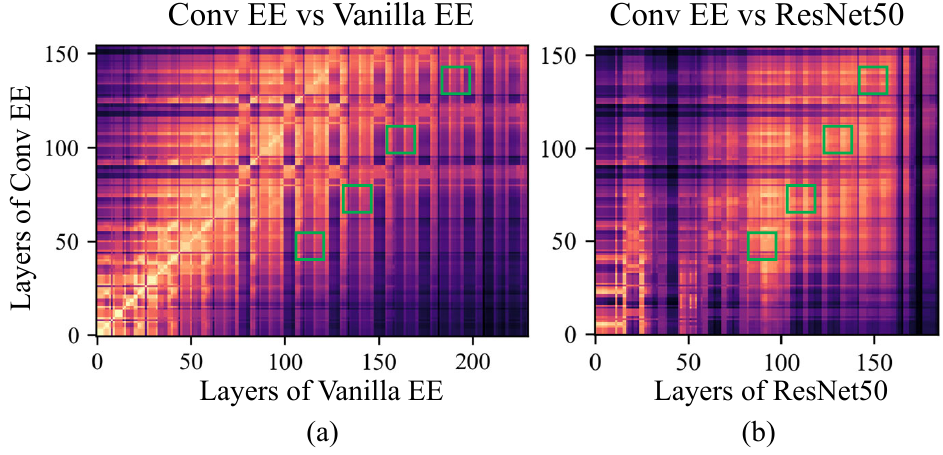}
  \end{center}
  \vspace{-9pt}
  \caption{CKA heatmap comparing vanilla EE vs Conv EE and Conv EE vs ResNet50. 
  Conv EE, which utilizes convolution as the exiting head, can learn different feature representations compared to vanilla EE. 
  The layer incorporating convolution in the internal classifier closely resembles the lower half of the ResNet layers. 
  Thus, Conv EE is more effective in capturing feature representations than vanilla EE.} 
  \Description{cka.}
  \label{fig:cka}
  \vspace{-6pt}
\end{figure}

\subsection{Outline}
In this supplementary material, we present a systematic investigation into the effectiveness of early exiting in ViTs. 
Besides, we provide more implementation details and experimental comparisons.
The main content is summarized as follows:
\begin{itemize}
  \item In Appendix \ref{sc:investigation}, we conduct a systematic investigation into the effectiveness of early exiting in ViTs and analyze the issues arising from the vanilla early exiting. 
  Moreover, we obtain two observations: \textit{1)} shallow internal classifiers cannot learn sufficient feature representation; \textit{2)} deep internal classifiers cannot capture target semantic information. 
  \item In Appendix \ref{sc:append_ex}, we perform additional experiments to evaluate our framework. We measure the actual execution time of different methods and compare four representative training schemes with different widely-used exiting heads. 
  Then, we ablate the effect of different exiting position schemes.
\end{itemize}

\subsection{Investigation of Early Exiting in ViT}
\label{sc:investigation}

The early exiting method can halt the forward propagation of neural networks prematurely to provide a speed-accuracy trade-off, 
which has achieved significant performance improvements for CNNs and transformers in NLP. 
However, naively implementing early exiting on ViT may not yield performance gains for internal classifiers. 
For instance, the performance of the internal classifier on the fifth and tenth layers decreases by 21.8\% and 4.0\%, respectively, 
compared to the original classifier for ViT-B\cite{vit} on CIFAR-100\cite{cifar}. 
Upon examining recent studies on ViT, we discover that a line of works focus on the integration of convolution and self-attention, 
which demonstrates that convolution operations at shallow layers can introduce additional inductive biases and capture more local information \cite{vit,acmix}.
Another line of works strive to exclusively employ self-attention as basic modules to construct the backbone with numerous layers for various vision tasks 
due to its exceptional capability in handling long-range dependencies \cite{scale_vit,deeper_vit,how_vit_work}. 
However, it still remains unexplored that 
\textit{1)} whether shallow internal classifiers could learn sufficient feature representation 
\textit{2)} and whether deep internal classifiers could capture target semantic information. 
Therefore, we design the following probing experiments to answer these two questions and systematically analyze the working mechanism of early exiting methods in ViT.

\begin{figure}[h]
  \vspace{-5pt}
  \begin{center}
    \includegraphics[width=0.45\textwidth]{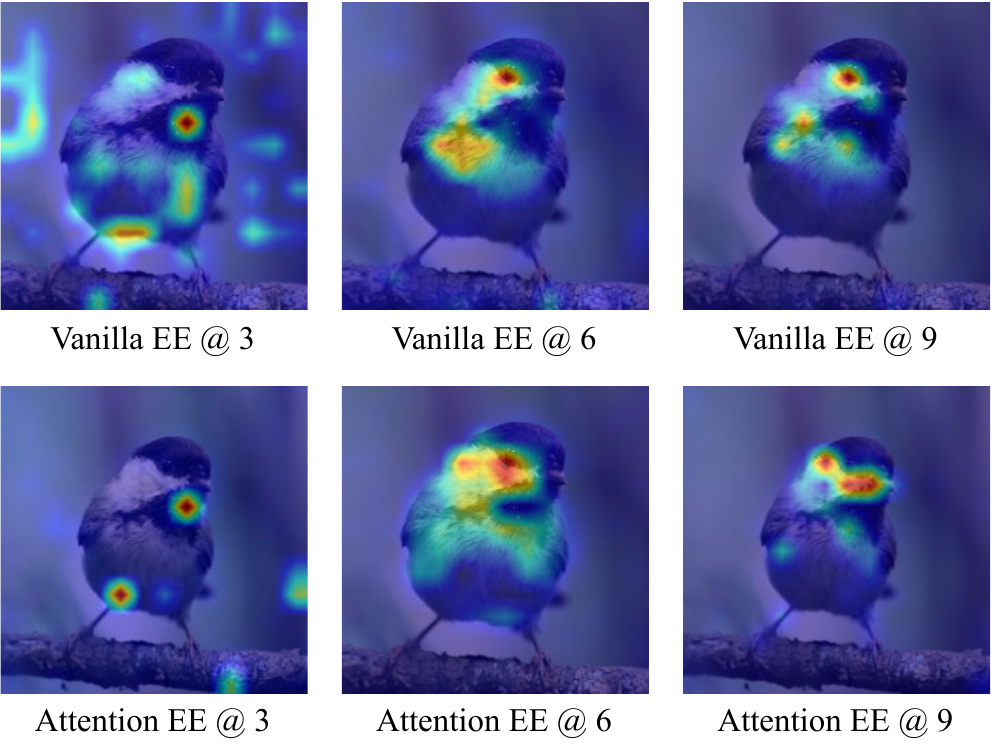}
  \end{center}
  \vspace{-5pt}
  \caption{
    Comparison of attention maps for early exiting in ViT using either MLP (vanilla EE) or self-attention (Attention EE) as the exiting head. 
    Attention EE methods are more capable of learning target semantic information by incorporating self-attention on deep internal classifiers than vanilla EE.}
  \Description{Attention map.}
  \label{fig:attention}
  \vspace{-8pt}
\end{figure}

Regrading the first question, we initially compare the representation similarity of two early exiting (EE) architectures, namely MLP (vanilla EE) and convolution (Conv EE), 
and subsequently assess the similarity between Conv EE and ResNet. 
We employ centered kernel alignment (CKA) \cite{cka} as the similarity metric, which facilitates quantitative comparisons of representations similarities within and across neural networks.
It is important to note that we compare their representation similarities using outputs from all internal classifiers and intermediate layers. 
The results are evaluated on the ViT-B/16 backbone using CIFAR-100 \cite{cifar}. 
We only utilize standard $3\times3$ convolution in Conv EE for fair comparison.
Figure \ref{fig:cka} displays the CKA similarity results as heatmaps, with the x and y axes indexing the layers from input to output. 
The layers attached MLP or convolution are marked with green boxes. 
Lighter colors in the heatmap indicate a higher representation similarity between the corresponding layers. 
We observe that the layer at which the internal classifier is attached in vanilla EE differs from that in Conv EE, 
suggesting that they extract distinct information, as depicted in Figure \ref{fig:cka} (a).
The layer where the convolution exiting architecture is positioned exhibits high similarity to ResNet, as shown in Figure \ref{fig:cka} (b).
Consequently, the convolution exiting architecture assists ViT in capturing local information and strengthening feature representations, 
allowing Conv EE to learn representations akin to those of ResNet.

Concerning the second question, we compare the ability to extract target semantic information between two different early exiting architectures, 
namely vanilla EE and self-attention (attention EE) at different exit positions. 
We compute the attention map and visualize it upon the input image following \cite{visualize}. 
The attention map highlights target pixels of the image that contribute to the dominance of the predicted label, 
enabling the analysis of the efficacy of semantic information extraction from the input space.
We employ DeiT-B \cite{deit} as the backbone, which comprises twelve layers in total. 
The attention map results for the third, sixth, and ninth layers are presented in Figure \ref{fig:attention}. 
It becomes more evident that the eye and beak of the bird are target parts for object identification using the attention EE method than the vanilla EE method, 
especially on the deeper layer, such as the ninth layer.
Since the attention EE method employs self-attention as the exiting architecture, the ability to extract semantic features and spatial relationships can be further enhanced. 
As a result, incorporating self-attention on deep layers can help learn more semantic representations and capture richer global information than the vanilla EE method.

Based on the aforementioned analyses, we obtain the following observations: 
\begin{itemize}
    \item \textbf{\textit{Observation 1}:} Shallow internal classifiers cannot learn sufficient feature representation. 
    \item \textbf{\textit{Observation 2}:} Deep internal classifiers cannot capture target semantic information. 
  \end{itemize}

\textbf{Insight.} 
We identify the primary reason for the poor performance resulting from directly applying the early exiting strategy in ViT. 
Furthermore, we discover that integrating convolution on shallow internal classifiers can enhance local information exploration, 
while incorporating self-attention on deep layers can improve the ability to obtain global information. 
Consequently, if both convolution and self-attention are employed as exiting architectures, positioned on shallow and deep layers respectively, 
the model would gain access to a more comprehensive combination of local and global information compared to using only MLP as the exiting architecture.

\subsection{Additional Experiments}
\label{sc:append_ex}

In this section, we evaluate the actual execution time of our method and compare four representative training schemes with different widely-used exiting heads. 
Moreover, we ablate the influences of different exiting position schemes. 

\textbf{Execution time.} We design experiments to measure the actual execution time with batch 1 on a RTX 3090 GPU. 
For each method, we run it for once as a warm-up and then record the execution time with 50 runs without break for the whole testing set of CIFAR-100. 
The results are shown in Table \ref{tb:time}. We can find that our method achieves the highest accuracy, the lowest running time and lowest computation cost. 

\begin{table}[!t]
  \centering
  \setlength{\extrarowheight}{0pt}
  \addtolength{\extrarowheight}{\aboverulesep}
  \addtolength{\extrarowheight}{\belowrulesep}
  \setlength{\aboverulesep}{0pt}
  \setlength{\belowrulesep}{0pt}
  \caption{Comparison of execution time on a RTX 3090 GPU.}
  \label{tb:time}
  \vspace{-6pt}
  \begin{tabular}{cccc} 
  \toprule
  \textbf{Method}                                 & \textbf{MACs}   & \textbf{Acc.}     & \textbf{Execution time}         \\ 
  \hline
  \textbf{ViT-B}                                  & 16.93 G         & 90.8 \%           & 6.49 ($\pm0.11$) ms            \\
  SDN                                             & 10.16 G         & 86.5 \%           & 5.65 ($\pm0.23$) ms            \\
  PABEE                                           & 11.48 G         & 85.1 \%           & 5.86 ($\pm0.09$) ms            \\
  PCEE                                            & 10.90 G         & 86.1 \%           & 6.70 ($\pm0.24$) ms            \\
  BERxiT                                          & 10.27 G         & 87.1 \%           & 5.75 ($\pm0.13$) ms~           \\
  ViT-EE                                          & 11.65 G         & 87.5 \%          & 5.35 ($\pm0.14$) ms~           \\
  \rowcolor[rgb]{0.949,0.949,0.949} \textbf{Ours} & \textbf{9.76 G} & \textbf{88.5} \% & \textbf{5.03} ($\pm0.16$) ms~  \\
  \bottomrule
  \end{tabular}
  \vspace{-5pt}
  \end{table}

  \begin{table}[t]
    \centering
    \setlength{\extrarowheight}{0pt}
    \addtolength{\extrarowheight}{\aboverulesep}
    \addtolength{\extrarowheight}{\belowrulesep}
    \setlength{\aboverulesep}{0pt}
    \setlength{\belowrulesep}{0pt}
    \caption{Comparison of different training schemes with widely-used exiting heads.}
    \label{tb:train_head}
    \vspace{-6pt}
    \begin{tabular}{cccc} 
    \toprule
    \textbf{Exiting heads}                      & \textbf{Training schemes} & \textbf{Acc.}     & \textbf{Speed-Up}       \\ 
    \hline
    MLP                                         & Normal                    & 87.3 \%           & 1.56~$\times$           \\
    MLP                                         & Weighted                  & 88.2 \%           & 1.53~$\times$~          \\
    MLP                                         & Distillation              & 87.1 \%           & 1.57~$\times$           \\
    MLP                                         & Alternating               & ~88.1 \%          & 1.54~$\times$~          \\
    \rowcolor[rgb]{0.949,0.949,0.949} MLP       & \textbf{Ours}             & ~\textbf{88.2}~\% & \textbf{1.57~}$\times$  \\ 
    \hline
    Conv                                        & Normal                    & 85.2 \%           & 1.69~$\times$           \\
    Conv                                        & Weighted                  & 86.4 \%           & 1.65~$\times$~          \\
    Conv                                        & Distillation              & 84.9 \%           & 1.64~$\times$           \\
    Conv                                        & Alternating               & ~86.7 \%          & 1.67~$\times$~          \\
    \rowcolor[rgb]{0.949,0.949,0.949} Conv      & \textbf{Ours}             & ~\textbf{87.5}~\% & \textbf{1.72~}$\times$  \\ 
    \hline
    Attention                                   & Normal                    & 86.8 \%           & 1.54~$\times$           \\
    Attention                                   & Weighted                  & 87.8 \%           & 1.53~$\times$~          \\
    Attention                                   & Distillation              & 87.0 \%           & 1.57~$\times$           \\
    Attention                                   & Alternating               & ~87.5 \%          & 1.54~$\times$~          \\
    \rowcolor[rgb]{0.949,0.949,0.949} Attention & \textbf{Ours}             & ~\textbf{88.0}~\% & \textbf{1.58~}$\times$  \\ 
    \hline
    Ours                                        & Normal                    & 86.9 \%           & 1.82~$\times$           \\
    Ours                                        & Weighted                  & 87.4 \%           & 1.80~$\times$~          \\
    Ours                                        & Distillation              & 86.9 \%           & 1.83~$\times$           \\
    Ours                                        & Alternating               & ~87.9 \%          & 1.84~$\times$~          \\
    \rowcolor[rgb]{0.949,0.949,0.949} Ours      & \textbf{Ours}             & ~\textbf{88.5}~\% & \textbf{1.87~}$\times$  \\
    \bottomrule
    \end{tabular}
    \vspace{-5pt}
    \end{table}

\textbf{Exiting heads \& training schemes.} We compare four representative training schemes with different widely-used exiting heads for the ViT backbone on CIFAR-100. 
The results of different training schemes with MLP, Conv, Attention and the proposed exiting heads are shown in Table \ref{tb:train_head}. 
We observe that our training scheme can improve classification accuracy and accelerate inference speed with different exiting heads.

  \begin{table}[h]
    \centering
    \setlength{\extrarowheight}{0pt}
    \addtolength{\extrarowheight}{\aboverulesep}
    \addtolength{\extrarowheight}{\belowrulesep}
    \setlength{\aboverulesep}{0pt}
    \setlength{\belowrulesep}{0pt}
    \caption{Comparison of different exiting positions on CIFAR-100. }
    \label{tb:position}
    \vspace{-6pt}
    \begin{tabular}{cccc} 
    \toprule
    \textbf{Position scheme}                                  & \textbf{Exiting position} & \textbf{Acc.}     & \textbf{Speed-Up}       \\ 
    \hline
    Shallow                                                   & \{2,3,4,5\}               & 85.3 \%           & 1.76~$\times$           \\
    Deep                                                      & \{8,9,10,11\}             & 87.9 \%           & 1.46~$\times$~          \\
    Middle                                                    & \{6,7,8,9\}               & 87.5 \%          & 1.74~$\times$~          \\
    \rowcolor[rgb]{0.949,0.949,0.949} Uniform (\textbf{Ours}) & \{4,6,8,10\}              & \textbf{88.4}~\% & \textbf{1.81~}$\times$  \\
    \bottomrule
    \end{tabular}
    \vspace{-2pt}
    \end{table}

\textbf{Exiting position.} In this work, all exiting heads are positioned across all layers uniformly according to an approximately equidistant computational distribution, i.e., 
the multiply-accumulate operations (MACs) of intermediate blocks between two adjacent exiting points remain consistent. 
We carefully ablate the influences of different exiting position schemes, including positioning at shallow, deep, and middle layers. 
The results of different exiting positions for the ViT backbone on CIFAR-100 dataset are shown in Table \ref{tb:position}.

%% file: main.bbl

\begin{thebibliography}{38}


\ifx \showCODEN    \undefined \def \showCODEN     #1{\unskip}     \fi
\ifx \showDOI      \undefined \def \showDOI       #1{#1}\fi
\ifx \showISBNx    \undefined \def \showISBNx     #1{\unskip}     \fi
\ifx \showISBNxiii \undefined \def \showISBNxiii  #1{\unskip}     \fi
\ifx \showISSN     \undefined \def \showISSN      #1{\unskip}     \fi
\ifx \showLCCN     \undefined \def \showLCCN      #1{\unskip}     \fi
\ifx \shownote     \undefined \def \shownote      #1{#1}          \fi
\ifx \showarticletitle \undefined \def \showarticletitle #1{#1}   \fi
\ifx \showURL      \undefined \def \showURL       {\relax}        \fi
\providecommand\bibfield[2]{#2}
\providecommand\bibinfo[2]{#2}
\providecommand\natexlab[1]{#1}
\providecommand\showeprint[2][]{arXiv:#2}

\bibitem[Bakhtiarnia et~al\mbox{.}(2021)]%
        {ViT-EE}
\bibfield{author}{\bibinfo{person}{Arian Bakhtiarnia}, \bibinfo{person}{Qi
  Zhang}, {and} \bibinfo{person}{Alexandros Iosifidis}.}
  \bibinfo{year}{2021}\natexlab{}.
\newblock \showarticletitle{Multi-Exit Vision Transformer for Dynamic
  Inference}. In \bibinfo{booktitle}{\emph{32nd British Machine Vision
  Conference 2021, {BMVC} 2021, Online, November 22-25, 2021}}.
  \bibinfo{publisher}{{BMVA} Press}, \bibinfo{pages}{81}.
\newblock


\bibitem[Bossard et~al\mbox{.}(2014)]%
        {food101}
\bibfield{author}{\bibinfo{person}{Lukas Bossard}, \bibinfo{person}{Matthieu
  Guillaumin}, {and} \bibinfo{person}{Luc~Van Gool}.}
  \bibinfo{year}{2014}\natexlab{}.
\newblock \showarticletitle{Food-101 - Mining Discriminative Components with
  Random Forests}. In \bibinfo{booktitle}{\emph{Computer Vision - {ECCV} 2014 -
  13th European Conference, Zurich, Switzerland, September 6-12, 2014,
  Proceedings, Part {VI}}} \emph{(\bibinfo{series}{Lecture Notes in Computer
  Science}, Vol.~\bibinfo{volume}{8694})},
  \bibfield{editor}{\bibinfo{person}{David~J. Fleet},
  \bibinfo{person}{Tom{\'{a}}s Pajdla}, \bibinfo{person}{Bernt Schiele}, {and}
  \bibinfo{person}{Tinne Tuytelaars}} (Eds.). \bibinfo{publisher}{Springer},
  \bibinfo{pages}{446--461}.
\newblock


\bibitem[Chefer et~al\mbox{.}(2021)]%
        {visualize}
\bibfield{author}{\bibinfo{person}{Hila Chefer}, \bibinfo{person}{Shir Gur},
  {and} \bibinfo{person}{Lior Wolf}.} \bibinfo{year}{2021}\natexlab{}.
\newblock \showarticletitle{Transformer Interpretability Beyond Attention
  Visualization}. In \bibinfo{booktitle}{\emph{{IEEE} Conference on Computer
  Vision and Pattern Recognition, {CVPR} 2021, virtual, June 19-25, 2021}}.
  \bibinfo{publisher}{Computer Vision Foundation / {IEEE}},
  \bibinfo{pages}{782--791}.
\newblock


\bibitem[Chen et~al\mbox{.}(2022)]%
        {scale_vit}
\bibfield{author}{\bibinfo{person}{Richard~J. Chen}, \bibinfo{person}{Chengkuan
  Chen}, \bibinfo{person}{Yicong Li}, \bibinfo{person}{Tiffany~Y. Chen},
  \bibinfo{person}{Andrew~D. Trister}, \bibinfo{person}{Rahul~G. Krishnan},
  {and} \bibinfo{person}{Faisal Mahmood}.} \bibinfo{year}{2022}\natexlab{}.
\newblock \showarticletitle{Scaling Vision Transformers to Gigapixel Images via
  Hierarchical Self-Supervised Learning}. In
  \bibinfo{booktitle}{\emph{{IEEE/CVF} Conference on Computer Vision and
  Pattern Recognition, {CVPR} 2022, New Orleans, LA, USA, June 18-24, 2022}}.
  \bibinfo{publisher}{{IEEE}}, \bibinfo{pages}{16123--16134}.
\newblock


\bibitem[Deng et~al\mbox{.}(2009)]%
        {imagenet}
\bibfield{author}{\bibinfo{person}{Jia Deng}, \bibinfo{person}{Wei Dong},
  \bibinfo{person}{Richard Socher}, \bibinfo{person}{Li{-}Jia Li},
  \bibinfo{person}{Kai Li}, {and} \bibinfo{person}{Li Fei{-}Fei}.}
  \bibinfo{year}{2009}\natexlab{}.
\newblock \showarticletitle{ImageNet: {A} large-scale hierarchical image
  database}. In \bibinfo{booktitle}{\emph{2009 {IEEE} Computer Society
  Conference on Computer Vision and Pattern Recognition}} (Miami, Florida,
  {USA}). \bibinfo{publisher}{{IEEE} Computer Society},
  \bibinfo{pages}{248--255}.
\newblock


\bibitem[Ding et~al\mbox{.}(2022)]%
        {DBLP:conf/mm/DingQYCLWL22}
\bibfield{author}{\bibinfo{person}{Yifu Ding}, \bibinfo{person}{Haotong Qin},
  \bibinfo{person}{Qinghua Yan}, \bibinfo{person}{Zhenhua Chai},
  \bibinfo{person}{Junjie Liu}, \bibinfo{person}{Xiaolin Wei}, {and}
  \bibinfo{person}{Xianglong Liu}.} \bibinfo{year}{2022}\natexlab{}.
\newblock \showarticletitle{Towards Accurate Post-Training Quantization for
  Vision Transformer}. In \bibinfo{booktitle}{\emph{{MM} '22: The 30th {ACM}
  International Conference on Multimedia, Lisboa, Portugal, October 10 - 14,
  2022}}, \bibfield{editor}{\bibinfo{person}{Jo{\~{a}}o Magalh{\~{a}}es},
  \bibinfo{person}{Alberto~Del Bimbo}, \bibinfo{person}{Shin'ichi Satoh},
  \bibinfo{person}{Nicu Sebe}, \bibinfo{person}{Xavier Alameda{-}Pineda},
  \bibinfo{person}{Qin Jin}, \bibinfo{person}{Vincent Oria}, {and}
  \bibinfo{person}{Laura Toni}} (Eds.). \bibinfo{publisher}{{ACM}},
  \bibinfo{pages}{5380--5388}.
\newblock


\bibitem[Dosovitskiy et~al\mbox{.}(2021)]%
        {vit}
\bibfield{author}{\bibinfo{person}{Alexey Dosovitskiy}, \bibinfo{person}{Lucas
  Beyer}, \bibinfo{person}{Alexander Kolesnikov}, \bibinfo{person}{Dirk
  Weissenborn}, \bibinfo{person}{Xiaohua Zhai}, \bibinfo{person}{Thomas
  Unterthiner}, \bibinfo{person}{Mostafa Dehghani}, \bibinfo{person}{Matthias
  Minderer}, \bibinfo{person}{Georg Heigold}, \bibinfo{person}{Sylvain Gelly},
  \bibinfo{person}{Jakob Uszkoreit}, {and} \bibinfo{person}{Neil Houlsby}.}
  \bibinfo{year}{2021}\natexlab{}.
\newblock \showarticletitle{An Image is Worth 16x16 Words: Transformers for
  Image Recognition at Scale}. In \bibinfo{booktitle}{\emph{9th International
  Conference on Learning Representations, {ICLR} 2021, Virtual Event, Austria,
  May 3-7, 2021}}. \bibinfo{publisher}{OpenReview.net}.
\newblock


\bibitem[Hao et~al\mbox{.}(2022)]%
        {hao2022learning}
\bibfield{author}{\bibinfo{person}{Zhiwei Hao}, \bibinfo{person}{Jianyuan Guo},
  \bibinfo{person}{Ding Jia}, \bibinfo{person}{Kai Han}, \bibinfo{person}{Yehui
  Tang}, \bibinfo{person}{Chao Zhang}, \bibinfo{person}{Han Hu}, {and}
  \bibinfo{person}{Yunhe Wang}.} \bibinfo{year}{2022}\natexlab{}.
\newblock \showarticletitle{Learning efficient vision transformers via
  fine-grained manifold distillation}.
\newblock \bibinfo{journal}{\emph{Advances in Neural Information Processing
  Systems}}  \bibinfo{volume}{35} (\bibinfo{year}{2022}),
  \bibinfo{pages}{9164--9175}.
\newblock


\bibitem[Kaya et~al\mbox{.}(2019)]%
        {SDN}
\bibfield{author}{\bibinfo{person}{Yigitcan Kaya}, \bibinfo{person}{Sanghyun
  Hong}, {and} \bibinfo{person}{Tudor Dumitras}.}
  \bibinfo{year}{2019}\natexlab{}.
\newblock \showarticletitle{Shallow-Deep Networks: Understanding and Mitigating
  Network Overthinking}. In \bibinfo{booktitle}{\emph{Proceedings of the 36th
  International Conference on Machine Learning, {ICML} 2019, 9-15 June 2019,
  Long Beach, California, {USA}}} \emph{(\bibinfo{series}{Proceedings of
  Machine Learning Research}, Vol.~\bibinfo{volume}{97})},
  \bibfield{editor}{\bibinfo{person}{Kamalika Chaudhuri} {and}
  \bibinfo{person}{Ruslan Salakhutdinov}} (Eds.). \bibinfo{publisher}{{PMLR}},
  \bibinfo{pages}{3301--3310}.
\newblock


\bibitem[Kornblith et~al\mbox{.}(2019)]%
        {cka}
\bibfield{author}{\bibinfo{person}{Simon Kornblith}, \bibinfo{person}{Mohammad
  Norouzi}, \bibinfo{person}{Honglak Lee}, {and} \bibinfo{person}{Geoffrey~E.
  Hinton}.} \bibinfo{year}{2019}\natexlab{}.
\newblock \showarticletitle{Similarity of Neural Network Representations
  Revisited}. In \bibinfo{booktitle}{\emph{Proceedings of the 36th
  International Conference on Machine Learning, {ICML} 2019, 9-15 June 2019,
  Long Beach, California, {USA}}} \emph{(\bibinfo{series}{Proceedings of
  Machine Learning Research}, Vol.~\bibinfo{volume}{97})},
  \bibfield{editor}{\bibinfo{person}{Kamalika Chaudhuri} {and}
  \bibinfo{person}{Ruslan Salakhutdinov}} (Eds.). \bibinfo{publisher}{{PMLR}},
  \bibinfo{pages}{3519--3529}.
\newblock


\bibitem[Krizhevsky and Hinton(2009)]%
        {cifar100}
\bibfield{author}{\bibinfo{person}{Alex Krizhevsky} {and}
  \bibinfo{person}{Geoffrey Hinton}.} \bibinfo{year}{2009}\natexlab{}.
\newblock \bibinfo{booktitle}{\emph{Learning multiple layers of features from
  tiny images}}.
\newblock \bibinfo{type}{{T}echnical {R}eport}. \bibinfo{address}{Toronto,
  Ontario}.
\newblock


\bibitem[Krizhevsky et~al\mbox{.}(2009)]%
        {cifar}
\bibfield{author}{\bibinfo{person}{Alex Krizhevsky}, \bibinfo{person}{Geoffrey
  Hinton}, {et~al\mbox{.}}} \bibinfo{year}{2009}\natexlab{}.
\newblock \showarticletitle{Learning multiple layers of features from tiny
  images}.
\newblock  (\bibinfo{year}{2009}).
\newblock


\bibitem[Kwon et~al\mbox{.}(2022)]%
        {kwon2022fast}
\bibfield{author}{\bibinfo{person}{Woosuk Kwon}, \bibinfo{person}{Sehoon Kim},
  \bibinfo{person}{Michael~W Mahoney}, \bibinfo{person}{Joseph Hassoun},
  \bibinfo{person}{Kurt Keutzer}, {and} \bibinfo{person}{Amir Gholami}.}
  \bibinfo{year}{2022}\natexlab{}.
\newblock \showarticletitle{A Fast Post-Training Pruning Framework for
  Transformers}.
\newblock \bibinfo{journal}{\emph{arXiv preprint arXiv:2204.09656}}
  (\bibinfo{year}{2022}).
\newblock


\bibitem[Liu et~al\mbox{.}(2020)]%
        {fastbert}
\bibfield{author}{\bibinfo{person}{Weijie Liu}, \bibinfo{person}{Peng Zhou},
  \bibinfo{person}{Zhiruo Wang}, \bibinfo{person}{Zhe Zhao},
  \bibinfo{person}{Haotang Deng}, {and} \bibinfo{person}{Qi Ju}.}
  \bibinfo{year}{2020}\natexlab{}.
\newblock \showarticletitle{FastBERT: a Self-distilling {BERT} with Adaptive
  Inference Time}. In \bibinfo{booktitle}{\emph{Proceedings of the 58th Annual
  Meeting of the Association for Computational Linguistics, {ACL} 2020, Online,
  July 5-10, 2020}}, \bibfield{editor}{\bibinfo{person}{Dan Jurafsky},
  \bibinfo{person}{Joyce Chai}, \bibinfo{person}{Natalie Schluter}, {and}
  \bibinfo{person}{Joel~R. Tetreault}} (Eds.). \bibinfo{publisher}{Association
  for Computational Linguistics}, \bibinfo{pages}{6035--6044}.
\newblock


\bibitem[Liu et~al\mbox{.}(2021)]%
        {swin}
\bibfield{author}{\bibinfo{person}{Ze Liu}, \bibinfo{person}{Yutong Lin},
  \bibinfo{person}{Yue Cao}, \bibinfo{person}{Han Hu}, \bibinfo{person}{Yixuan
  Wei}, \bibinfo{person}{Zheng Zhang}, \bibinfo{person}{Stephen Lin}, {and}
  \bibinfo{person}{Baining Guo}.} \bibinfo{year}{2021}\natexlab{}.
\newblock \showarticletitle{Swin Transformer: Hierarchical Vision Transformer
  using Shifted Windows}. In \bibinfo{booktitle}{\emph{2021 {IEEE/CVF}
  International Conference on Computer Vision, {ICCV} 2021, Montreal, QC,
  Canada, October 10-17, 2021}}. \bibinfo{publisher}{{IEEE}},
  \bibinfo{pages}{9992--10002}.
\newblock


\bibitem[Maaz et~al\mbox{.}(2022)]%
        {DBLP:conf/eccv/MaazSCKZAK22}
\bibfield{author}{\bibinfo{person}{Muhammad Maaz}, \bibinfo{person}{Abdelrahman
  Shaker}, \bibinfo{person}{Hisham Cholakkal}, \bibinfo{person}{Salman~H.
  Khan}, \bibinfo{person}{Syed~Waqas Zamir}, \bibinfo{person}{Rao~Muhammad
  Anwer}, {and} \bibinfo{person}{Fahad~Shahbaz Khan}.}
  \bibinfo{year}{2022}\natexlab{}.
\newblock \showarticletitle{EdgeNeXt: Efficiently Amalgamated CNN-Transformer
  Architecture for Mobile Vision Applications}. In
  \bibinfo{booktitle}{\emph{Computer Vision - {ECCV} 2022 Workshops - Tel Aviv,
  Israel, October 23-27, 2022, Proceedings, Part {VII}}}
  \emph{(\bibinfo{series}{Lecture Notes in Computer Science},
  Vol.~\bibinfo{volume}{13807})}, \bibfield{editor}{\bibinfo{person}{Leonid
  Karlinsky}, \bibinfo{person}{Tomer Michaeli}, {and}
  \bibinfo{person}{Ko~Nishino}} (Eds.). \bibinfo{publisher}{Springer},
  \bibinfo{pages}{3--20}.
\newblock


\bibitem[Mehta and Rastegari(2022)]%
        {mobilevit}
\bibfield{author}{\bibinfo{person}{Sachin Mehta} {and}
  \bibinfo{person}{Mohammad Rastegari}.} \bibinfo{year}{2022}\natexlab{}.
\newblock \showarticletitle{MobileViT: Light-weight, General-purpose, and
  Mobile-friendly Vision Transformer}. In \bibinfo{booktitle}{\emph{The Tenth
  International Conference on Learning Representations, {ICLR} 2022, Virtual
  Event, April 25-29, 2022}}. \bibinfo{publisher}{OpenReview.net}.
\newblock


\bibitem[Pan et~al\mbox{.}(2022)]%
        {acmix}
\bibfield{author}{\bibinfo{person}{Xuran Pan}, \bibinfo{person}{Chunjiang Ge},
  \bibinfo{person}{Rui Lu}, \bibinfo{person}{Shiji Song},
  \bibinfo{person}{Guanfu Chen}, \bibinfo{person}{Zeyi Huang}, {and}
  \bibinfo{person}{Gao Huang}.} \bibinfo{year}{2022}\natexlab{}.
\newblock \showarticletitle{On the Integration of Self-Attention and
  Convolution}. In \bibinfo{booktitle}{\emph{{IEEE/CVF} Conference on Computer
  Vision and Pattern Recognition, {CVPR} 2022, New Orleans, LA, USA, June
  18-24, 2022}}. \bibinfo{publisher}{{IEEE}}, \bibinfo{pages}{805--815}.
\newblock


\bibitem[Park and Kim(2022)]%
        {how_vit_work}
\bibfield{author}{\bibinfo{person}{Namuk Park} {and} \bibinfo{person}{Songkuk
  Kim}.} \bibinfo{year}{2022}\natexlab{}.
\newblock \showarticletitle{How Do Vision Transformers Work?}. In
  \bibinfo{booktitle}{\emph{The Tenth International Conference on Learning
  Representations, {ICLR} 2022, Virtual Event, April 25-29, 2022}}.
  \bibinfo{publisher}{OpenReview.net}.
\newblock


\bibitem[Sajjad et~al\mbox{.}(2023)]%
        {DBLP:journals/csl/SajjadDDN23}
\bibfield{author}{\bibinfo{person}{Hassan Sajjad}, \bibinfo{person}{Fahim
  Dalvi}, \bibinfo{person}{Nadir Durrani}, {and} \bibinfo{person}{Preslav
  Nakov}.} \bibinfo{year}{2023}\natexlab{}.
\newblock \showarticletitle{On the effect of dropping layers of pre-trained
  transformer models}.
\newblock \bibinfo{journal}{\emph{Comput. Speech Lang.}}  \bibinfo{volume}{77}
  (\bibinfo{year}{2023}), \bibinfo{pages}{101429}.
\newblock


\bibitem[Shen et~al\mbox{.}(2023)]%
        {shen2023efficient}
\bibfield{author}{\bibinfo{person}{Li Shen}, \bibinfo{person}{Yan Sun},
  \bibinfo{person}{Zhiyuan Yu}, \bibinfo{person}{Liang Ding},
  \bibinfo{person}{Xinmei Tian}, {and} \bibinfo{person}{Dacheng Tao}.}
  \bibinfo{year}{2023}\natexlab{}.
\newblock \showarticletitle{On Efficient Training of Large-Scale Deep Learning
  Models: A Literature Review}.
\newblock \bibinfo{journal}{\emph{arXiv preprint arXiv:2304.03589}}
  (\bibinfo{year}{2023}).
\newblock


\bibitem[Sun et~al\mbox{.}(2022)]%
        {hash}
\bibfield{author}{\bibinfo{person}{Tianxiang Sun}, \bibinfo{person}{Xiangyang
  Liu}, \bibinfo{person}{Wei Zhu}, \bibinfo{person}{Zhichao Geng},
  \bibinfo{person}{Lingling Wu}, \bibinfo{person}{Yilong He},
  \bibinfo{person}{Yuan Ni}, \bibinfo{person}{Guotong Xie},
  \bibinfo{person}{Xuanjing Huang}, {and} \bibinfo{person}{Xipeng Qiu}.}
  \bibinfo{year}{2022}\natexlab{}.
\newblock \showarticletitle{A Simple Hash-Based Early Exiting Approach For
  Language Understanding and Generation}. In \bibinfo{booktitle}{\emph{Findings
  of the Association for Computational Linguistics: {ACL} 2022, Dublin,
  Ireland, May 22-27, 2022}}, \bibfield{editor}{\bibinfo{person}{Smaranda
  Muresan}, \bibinfo{person}{Preslav Nakov}, {and} \bibinfo{person}{Aline
  Villavicencio}} (Eds.). \bibinfo{publisher}{Association for Computational
  Linguistics}, \bibinfo{pages}{2409--2421}.
\newblock


\bibitem[Tang et~al\mbox{.}(2022)]%
        {DBLP:conf/cvpr/Tang00XGXT22}
\bibfield{author}{\bibinfo{person}{Yehui Tang}, \bibinfo{person}{Kai Han},
  \bibinfo{person}{Yunhe Wang}, \bibinfo{person}{Chang Xu},
  \bibinfo{person}{Jianyuan Guo}, \bibinfo{person}{Chao Xu}, {and}
  \bibinfo{person}{Dacheng Tao}.} \bibinfo{year}{2022}\natexlab{}.
\newblock \showarticletitle{Patch Slimming for Efficient Vision Transformers}.
  In \bibinfo{booktitle}{\emph{{IEEE/CVF} Conference on Computer Vision and
  Pattern Recognition, {CVPR} 2022, New Orleans, LA, USA, June 18-24, 2022}}.
  \bibinfo{publisher}{{IEEE}}, \bibinfo{pages}{12155--12164}.
\newblock


\bibitem[Tay et~al\mbox{.}(2023)]%
        {DBLP:journals/csur/TayDBM23}
\bibfield{author}{\bibinfo{person}{Yi Tay}, \bibinfo{person}{Mostafa Dehghani},
  \bibinfo{person}{Dara Bahri}, {and} \bibinfo{person}{Donald Metzler}.}
  \bibinfo{year}{2023}\natexlab{}.
\newblock \showarticletitle{Efficient Transformers: {A} Survey}.
\newblock \bibinfo{journal}{\emph{{ACM} Comput. Surv.}} \bibinfo{volume}{55},
  \bibinfo{number}{6} (\bibinfo{year}{2023}), \bibinfo{pages}{109:1--109:28}.
\newblock


\bibitem[Teerapittayanon et~al\mbox{.}(2016)]%
        {branchynet}
\bibfield{author}{\bibinfo{person}{Surat Teerapittayanon},
  \bibinfo{person}{Bradley McDanel}, {and} \bibinfo{person}{H.~T. Kung}.}
  \bibinfo{year}{2016}\natexlab{}.
\newblock \showarticletitle{BranchyNet: Fast inference via early exiting from
  deep neural networks}. In \bibinfo{booktitle}{\emph{23rd International
  Conference on Pattern Recognition, {ICPR} 2016, Canc{\'{u}}n, Mexico,
  December 4-8, 2016}}. \bibinfo{publisher}{{IEEE}},
  \bibinfo{pages}{2464--2469}.
\newblock


\bibitem[Touvron et~al\mbox{.}(2021a)]%
        {deit}
\bibfield{author}{\bibinfo{person}{Hugo Touvron}, \bibinfo{person}{Matthieu
  Cord}, \bibinfo{person}{Matthijs Douze}, \bibinfo{person}{Francisco Massa},
  \bibinfo{person}{Alexandre Sablayrolles}, {and} \bibinfo{person}{Herv{\'{e}}
  J{\'{e}}gou}.} \bibinfo{year}{2021}\natexlab{a}.
\newblock \showarticletitle{Training data-efficient image transformers {\&}
  distillation through attention}. In \bibinfo{booktitle}{\emph{Proceedings of
  the 38th International Conference on Machine Learning, {ICML} 2021, 18-24
  July 2021, Virtual Event}} \emph{(\bibinfo{series}{Proceedings of Machine
  Learning Research}, Vol.~\bibinfo{volume}{139})},
  \bibfield{editor}{\bibinfo{person}{Marina Meila} {and} \bibinfo{person}{Tong
  Zhang}} (Eds.). \bibinfo{publisher}{{PMLR}}, \bibinfo{pages}{10347--10357}.
\newblock


\bibitem[Touvron et~al\mbox{.}(2021b)]%
        {deeper_vit}
\bibfield{author}{\bibinfo{person}{Hugo Touvron}, \bibinfo{person}{Matthieu
  Cord}, \bibinfo{person}{Alexandre Sablayrolles}, \bibinfo{person}{Gabriel
  Synnaeve}, {and} \bibinfo{person}{Herv{\'{e}} J{\'{e}}gou}.}
  \bibinfo{year}{2021}\natexlab{b}.
\newblock \showarticletitle{Going deeper with Image Transformers}. In
  \bibinfo{booktitle}{\emph{2021 {IEEE/CVF} International Conference on
  Computer Vision, {ICCV} 2021, Montreal, QC, Canada, October 10-17, 2021}}.
  \bibinfo{publisher}{{IEEE}}, \bibinfo{pages}{32--42}.
\newblock


\bibitem[Wolczyk et~al\mbox{.}(2021)]%
        {ztw}
\bibfield{author}{\bibinfo{person}{Maciej Wolczyk}, \bibinfo{person}{Bartosz
  W{\'{o}}jcik}, \bibinfo{person}{Klaudia Balazy}, \bibinfo{person}{Igor~T.
  Podolak}, \bibinfo{person}{Jacek Tabor}, \bibinfo{person}{Marek Smieja},
  {and} \bibinfo{person}{Tomasz Trzcinski}.} \bibinfo{year}{2021}\natexlab{}.
\newblock \showarticletitle{Zero Time Waste: Recycling Predictions in Early
  Exit Neural Networks}. In \bibinfo{booktitle}{\emph{Advances in Neural
  Information Processing Systems 34: Annual Conference on Neural Information
  Processing Systems 2021, NeurIPS 2021, December 6-14, 2021, virtual}},
  \bibfield{editor}{\bibinfo{person}{Marc'Aurelio Ranzato},
  \bibinfo{person}{Alina Beygelzimer}, \bibinfo{person}{Yann~N. Dauphin},
  \bibinfo{person}{Percy Liang}, {and} \bibinfo{person}{Jennifer~Wortman
  Vaughan}} (Eds.). \bibinfo{pages}{2516--2528}.
\newblock


\bibitem[Wolf et~al\mbox{.}(2020)]%
        {huggingface}
\bibfield{author}{\bibinfo{person}{Thomas Wolf}, \bibinfo{person}{Lysandre
  Debut}, \bibinfo{person}{Victor Sanh}, \bibinfo{person}{Julien Chaumond},
  \bibinfo{person}{Clement Delangue}, \bibinfo{person}{Anthony Moi},
  \bibinfo{person}{Pierric Cistac}, \bibinfo{person}{Tim Rault},
  \bibinfo{person}{Rémi Louf}, \bibinfo{person}{Morgan Funtowicz},
  \bibinfo{person}{Joe Davison}, \bibinfo{person}{Sam Shleifer},
  \bibinfo{person}{Patrick von Platen}, \bibinfo{person}{Clara Ma},
  \bibinfo{person}{Yacine Jernite}, \bibinfo{person}{Julien Plu},
  \bibinfo{person}{Canwen Xu}, \bibinfo{person}{Teven~Le Scao},
  \bibinfo{person}{Sylvain Gugger}, \bibinfo{person}{Mariama Drame},
  \bibinfo{person}{Quentin Lhoest}, {and} \bibinfo{person}{Alexander~M. Rush}.}
  \bibinfo{year}{2020}\natexlab{}.
\newblock \showarticletitle{Transformers: State-of-the-Art Natural Language
  Processing}. In \bibinfo{booktitle}{\emph{Proceedings of the 2020 Conference
  on Empirical Methods in Natural Language Processing: System Demonstrations}}.
  \bibinfo{publisher}{Association for Computational Linguistics},
  \bibinfo{address}{Online}, \bibinfo{pages}{38--45}.
\newblock
\urldef\tempurl%
\url{https://www.aclweb.org/anthology/2020.emnlp-demos.6}
\showURL{%
\tempurl}


\bibitem[Wu et~al\mbox{.}(2022)]%
        {DBLP:conf/mm/WuRWW022}
\bibfield{author}{\bibinfo{person}{Zhenyu Wu}, \bibinfo{person}{Zhou Ren},
  \bibinfo{person}{Yi Wu}, \bibinfo{person}{Zhangyang Wang}, {and}
  \bibinfo{person}{Gang Hua}.} \bibinfo{year}{2022}\natexlab{}.
\newblock \showarticletitle{TxVAD: Improved Video Action Detection by
  Transformers}. In \bibinfo{booktitle}{\emph{{MM} '22: The 30th {ACM}
  International Conference on Multimedia, Lisboa, Portugal, October 10 - 14,
  2022}}, \bibfield{editor}{\bibinfo{person}{Jo{\~{a}}o Magalh{\~{a}}es},
  \bibinfo{person}{Alberto~Del Bimbo}, \bibinfo{person}{Shin'ichi Satoh},
  \bibinfo{person}{Nicu Sebe}, \bibinfo{person}{Xavier Alameda{-}Pineda},
  \bibinfo{person}{Qin Jin}, \bibinfo{person}{Vincent Oria}, {and}
  \bibinfo{person}{Laura Toni}} (Eds.). \bibinfo{publisher}{{ACM}},
  \bibinfo{pages}{4605--4613}.
\newblock


\bibitem[Xin et~al\mbox{.}(2020)]%
        {deebert}
\bibfield{author}{\bibinfo{person}{Ji Xin}, \bibinfo{person}{Raphael Tang},
  \bibinfo{person}{Jaejun Lee}, \bibinfo{person}{Yaoliang Yu}, {and}
  \bibinfo{person}{Jimmy Lin}.} \bibinfo{year}{2020}\natexlab{}.
\newblock \showarticletitle{DeeBERT: Dynamic Early Exiting for Accelerating
  {BERT} Inference}. In \bibinfo{booktitle}{\emph{Proceedings of the 58th
  Annual Meeting of the Association for Computational Linguistics, {ACL} 2020,
  Online, July 5-10, 2020}}, \bibfield{editor}{\bibinfo{person}{Dan Jurafsky},
  \bibinfo{person}{Joyce Chai}, \bibinfo{person}{Natalie Schluter}, {and}
  \bibinfo{person}{Joel~R. Tetreault}} (Eds.). \bibinfo{publisher}{Association
  for Computational Linguistics}, \bibinfo{pages}{2246--2251}.
\newblock


\bibitem[Xin et~al\mbox{.}(2021)]%
        {BERxiT}
\bibfield{author}{\bibinfo{person}{Ji Xin}, \bibinfo{person}{Raphael Tang},
  \bibinfo{person}{Yaoliang Yu}, {and} \bibinfo{person}{Jimmy Lin}.}
  \bibinfo{year}{2021}\natexlab{}.
\newblock \showarticletitle{BERxiT: Early Exiting for {BERT} with Better
  Fine-Tuning and Extension to Regression}. In
  \bibinfo{booktitle}{\emph{Proceedings of the 16th Conference of the European
  Chapter of the Association for Computational Linguistics: Main Volume, {EACL}
  2021, Online, April 19 - 23, 2021}}. \bibinfo{publisher}{Association for
  Computational Linguistics}, \bibinfo{pages}{91--104}.
\newblock


\bibitem[Yang et~al\mbox{.}(2022)]%
        {tvformer}
\bibfield{author}{\bibinfo{person}{Li Yang}, \bibinfo{person}{Mai Xu},
  \bibinfo{person}{Tie Liu}, \bibinfo{person}{Liangyu Huo}, {and}
  \bibinfo{person}{Xinbo Gao}.} \bibinfo{year}{2022}\natexlab{}.
\newblock \showarticletitle{TVFormer: Trajectory-guided Visual Quality
  Assessment on 360{\textdegree} Images with Transformers}. In
  \bibinfo{booktitle}{\emph{{MM} '22: The 30th {ACM} International Conference
  on Multimedia, Lisboa, Portugal, October 10 - 14, 2022}},
  \bibfield{editor}{\bibinfo{person}{Jo{\~{a}}o Magalh{\~{a}}es},
  \bibinfo{person}{Alberto~Del Bimbo}, \bibinfo{person}{Shin'ichi Satoh},
  \bibinfo{person}{Nicu Sebe}, \bibinfo{person}{Xavier Alameda{-}Pineda},
  \bibinfo{person}{Qin Jin}, \bibinfo{person}{Vincent Oria}, {and}
  \bibinfo{person}{Laura Toni}} (Eds.). \bibinfo{publisher}{{ACM}},
  \bibinfo{pages}{799--808}.
\newblock


\bibitem[Yuan et~al\mbox{.}(2022)]%
        {yuan2022ptq4vit}
\bibfield{author}{\bibinfo{person}{Zhihang Yuan}, \bibinfo{person}{Chenhao
  Xue}, \bibinfo{person}{Yiqi Chen}, \bibinfo{person}{Qiang Wu}, {and}
  \bibinfo{person}{Guangyu Sun}.} \bibinfo{year}{2022}\natexlab{}.
\newblock \showarticletitle{Ptq4vit: Post-training quantization for vision
  transformers with twin uniform quantization}. In
  \bibinfo{booktitle}{\emph{European Conference on Computer Vision}}. Springer,
  \bibinfo{pages}{191--207}.
\newblock


\bibitem[Zhang et~al\mbox{.}(2022a)]%
        {self-distill}
\bibfield{author}{\bibinfo{person}{Linfeng Zhang}, \bibinfo{person}{Chenglong
  Bao}, {and} \bibinfo{person}{Kaisheng Ma}.} \bibinfo{year}{2022}\natexlab{a}.
\newblock \showarticletitle{Self-Distillation: Towards Efficient and Compact
  Neural Networks}.
\newblock \bibinfo{journal}{\emph{{IEEE} Trans. Pattern Anal. Mach. Intell.}}
  \bibinfo{volume}{44}, \bibinfo{number}{8} (\bibinfo{year}{2022}),
  \bibinfo{pages}{4388--4403}.
\newblock


\bibitem[Zhang et~al\mbox{.}(2022b)]%
        {PCEE}
\bibfield{author}{\bibinfo{person}{Zhen Zhang}, \bibinfo{person}{Wei Zhu},
  \bibinfo{person}{Jinfan Zhang}, \bibinfo{person}{Peng Wang},
  \bibinfo{person}{Rize Jin}, {and} \bibinfo{person}{Tae{-}Sun Chung}.}
  \bibinfo{year}{2022}\natexlab{b}.
\newblock \showarticletitle{{PCEE-BERT:} Accelerating {BERT} Inference via
  Patient and Confident Early Exiting}. In \bibinfo{booktitle}{\emph{Findings
  of the Association for Computational Linguistics: {NAACL} 2022, Seattle, WA,
  United States, July 10-15, 2022}}. \bibinfo{publisher}{Association for
  Computational Linguistics}, \bibinfo{pages}{327--338}.
\newblock


\bibitem[Zheng et~al\mbox{.}(2022)]%
        {DBLP:conf/nips/ZhenglZYTXRP22}
\bibfield{author}{\bibinfo{person}{Chuanyang Zheng}, \bibinfo{person}{Zheyang
  Li}, \bibinfo{person}{Kai Zhang}, \bibinfo{person}{Zhi Yang},
  \bibinfo{person}{Wenming Tan}, \bibinfo{person}{Jun Xiao},
  \bibinfo{person}{Ye Ren}, {and} \bibinfo{person}{Shiliang Pu}.}
  \bibinfo{year}{2022}\natexlab{}.
\newblock \showarticletitle{SAViT: Structure-Aware Vision Transformer Pruning
  via Collaborative Optimization}. In \bibinfo{booktitle}{\emph{NeurIPS}}.
\newblock


\bibitem[Zhou et~al\mbox{.}(2020)]%
        {PABEE}
\bibfield{author}{\bibinfo{person}{Wangchunshu Zhou}, \bibinfo{person}{Canwen
  Xu}, \bibinfo{person}{Tao Ge}, \bibinfo{person}{Julian~J. McAuley},
  \bibinfo{person}{Ke Xu}, {and} \bibinfo{person}{Furu Wei}.}
  \bibinfo{year}{2020}\natexlab{}.
\newblock \showarticletitle{{BERT} Loses Patience: Fast and Robust Inference
  with Early Exit}. In \bibinfo{booktitle}{\emph{Advances in Neural Information
  Processing Systems 33: Annual Conference on Neural Information Processing
  Systems 2020, NeurIPS 2020, December 6-12, 2020, virtual}}.
\newblock


\end{thebibliography}
